\documentclass[10pt,twocolumn,letterpaper]{article}

\usepackage[pagenumbers]{cvpr} 

\definecolor{cvprblue}{rgb}{0.21,0.49,0.74}
\usepackage[pagebackref,breaklinks,colorlinks,allcolors=cvprblue]{hyperref}
\usepackage{multirow}
\usepackage{pifont}
\usepackage{cuted}
\usepackage[table]{xcolor}
\usepackage{xcolor}
\usepackage{tikz}
\usepackage[accsupp]{axessibility}  

\newcommand{\solidcell}[2]{
\tikz[baseline=(n.base)]{
  \node[
    inner sep=0.4ex,
    outer sep=0pt,
    minimum width=\linewidth, 
    text width=\linewidth,   
    align=left,                 
    anchor=west,               
    minimum height=1.6em,       
    fill=#1                 
  ] (n) {\strut #2};
}}

\definecolor{lightyellow}{rgb}{1,1, 0.8}
\definecolor{yellow}{rgb}{1,0.97, 0.65}
\definecolor{orange}{rgb}{1, 0.85, 0.7}
\definecolor{tablered}{rgb}{1, 0.7, 0.7}
\definecolor{tablegreen}{rgb}{0.80, 1, 0.80}
\definecolor{tablegreen2}{rgb}{0.90, 1, 0.90}
\definecolor{tablered2}{rgb}{0.8, 0.8, 1.0}

\def\paperID{} 
\def\confName{CVPR}
\def\confYear{2026}

\newcommand{\ours}{E-RayZer}
\newcommand{\ERayZerLogo}{%
  \mbox{%
    {\fontfamily{cmr}\fontseries{bx}\fontshape{it}\selectfont E}%
    \kern0.4pt%
    {\fontfamily{cmr}\fontseries{m}\fontshape{it}\selectfont -RayZer:}%
  }%
}
\newcommand{\unitA}{{\bfseries Self-supervised 3D Reconstruction}}
\newcommand{\unitB}{{\bfseries Spatial Visual Pre-training}}

\title{\ERayZerLogo \ 
  \unitA\ as\ \unitB}

\author{
Qitao Zhao$^{1}$~~~
Hao Tan$^{2}$~~~
Qianqian Wang$^{3}$~~~
Sai Bi$^{2}$~~~\\
Kai Zhang$^{2}$~~~
Kalyan Sunkavalli$^{2}$~~~
Shubham Tulsiani$^{1*}$~~~
Hanwen Jiang$^{2*}$~~~
\\
\small{
$^{1}$Carnegie Mellon University \ 
$^{2}$Adobe Research \
$^{3}$Harvard University \
$^{*}$Equal advising
}
\\
\small{
\textit{Project \& Code:} \href{https://qitaozhao.github.io/E-RayZer}{qitaozhao.github.io/E-RayZer}
}
}

\begin{document}
\maketitle
\begin{strip}
\vspace{-45pt}
\centering
\includegraphics[width=1.0\textwidth]{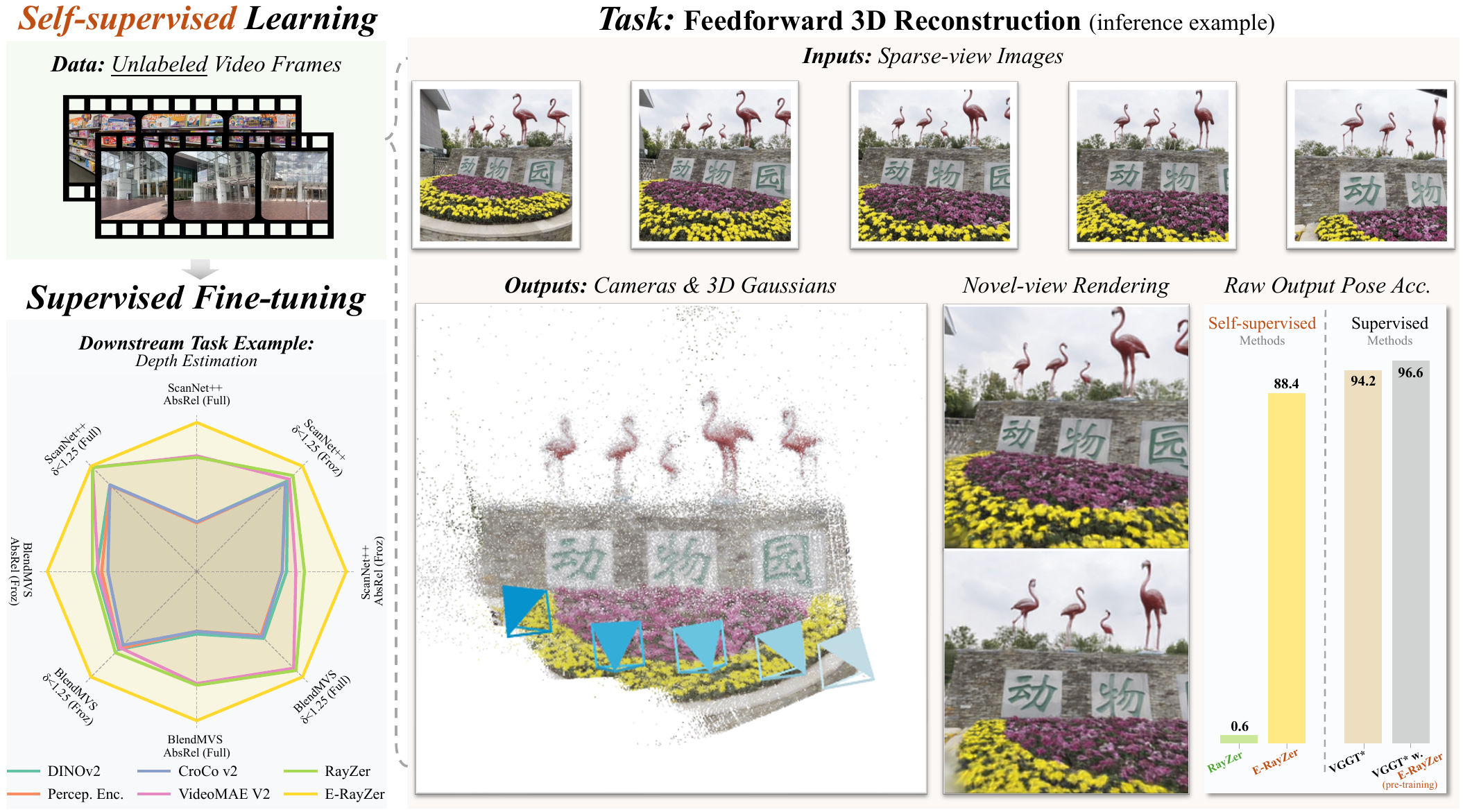}
\vspace{-0.3in}
\captionsetup{hypcap=false}\captionof{figure}{\textbf{\ours{}}, a \textbf{self-supervised} 3D vision model that predicts camera poses and scene geometry as 3D Gaussians. The use of \textbf{explicit} 3D geometry yields more geometrically grounded poses compared to its implicit counterpart, RayZer~\cite{jiang2025rayzer}: they are comparable to (and sometimes surpass) those from our supervised baseline, VGGT~\cite{wang2025vggt}. Furthermore, \ours{} serves as a self-supervised \textbf{visual pre-training} framework, with learned representations that transfer effectively to downstream tasks requiring 3D understanding, outperforming previous representation learners such as CroCo v2~\cite{weinzaepfel2023croco}, VideoMAE V2~\cite{wang2023videomae}, DINOv3~\cite{simeoni2025dinov3}, and Perception Encoder~\cite{bolya2025perception}.}
\vspace{-0.06in}
\label{fig:teaser}
\end{strip}

\begin{abstract}

\vspace{-0.19in}
Self-supervised pre-training has driven rapid progress in foundation models for language, 2D images, and video, yet remains largely unexplored for learning 3D-aware representations from multi-view images. In this paper, we present \ours{}, a self-supervised 3D vision model that learns geometrically grounded representations directly from unlabeled images. Unlike prior self-supervised methods such as  RayZer, which infer 3D indirectly through latent-space view synthesis, \textbf{E}-RayZer operates directly in 3D space, performing self-supervised 3D reconstruction with \textbf{Explicit} geometry. This formulation eliminates shortcut solutions and yields representations that are 3D-aware. To ensure convergence and scalability, we introduce a fine-grained learning curriculum that organizes training from easy to hard samples and harmonizes heterogeneous data sources without any supervision. Experiments show that \ours{} significantly outperforms RayZer on pose estimation and matches or sometimes surpasses fully supervised reconstruction models such as VGGT\footnote{We reproduce VGGT with matched model architecture and training setups, and denote it by VGGT* as a supervised baseline.}. Furthermore, its learned representations outperform leading visual pre-training models (\eg, DINOv3, CroCo v2, VideoMAE V2, and RayZer) on 3D downstream tasks, establishing \ours{} as a promising paradigm for spatial visual pre-training.

\end{abstract}    
\vspace{-0.2in}
\section{Introduction}

\textbf{Pre-training with self-supervision} forms the foundation of frontier models, enabling them to learn \textit{meaningful representations} from vast amounts of unlabeled data. This paradigm has proven effective for text~\cite{devlin2019bert, brown2020language}, 2D images~\cite{oquab2023dinov2, he2022masked}, and video~\cite{tong2022videomae, assran2025v}, where large models capture language semantics, visual concepts, and temporal dynamics. However, we argue that one essential component is still missing -- \textbf{learning 3D-aware representations from unlabeled multi-view images} -- as 3D spatial understanding is fundamental for perceiving and interacting with the physical world. Yet current 3D vision models mostly follow a different route: \textit{fully supervised learning} using 3D pseudo-labels estimated by SfM systems (\eg, COLMAP~\cite{schonberger2016structure}), which is inherently inefficient, imperfect, and ultimately unscalable. To move forward, we need a self-supervised pre-training framework that can learn 3D-aware representations from abundant raw visual observations.

In this paper, we present \textbf{\ours{}}, the first \textbf{truly self-supervised 3D Gaussian splatting reconstruction} model that learns 3D-aware representations from unlabeled data, establishing a new paradigm for \textit{3D spatial visual pre-training} (Fig.~\ref{fig:teaser}).
Unlike its predecessor RayZer~\cite{jiang2025rayzer}, which exhibits only \textit{superficial} 3D awareness through the \textit{proxy task} of self-supervised view synthesis in \textit{latent space}, \ours{} operates directly in \textit{3D space}, learning self-supervised 3D reconstruction.
Concretely, \ours{} predicts camera parameters and 3D Gaussians~\cite{kerbl20233d} from inputs and renders them back for photometric self-supervision under physical rendering constraints.
By grounding representations in \textbf{explicit} scene geometry, \ours{} learns features that are genuinely 3D-aware and free from RayZer's shortcut solutions such as frame interpolation (see Sec.~\ref{sec:preliminaries}).
This design yields both a \textit{camera space} that is more geometrically grounded and interpretable than RayZer's, and \textit{latent representations} that are truly \textbf{3D-aware}, effectively benefiting downstream 3D vision tasks.

Although explicit 3D Gaussians offer clear advantages, they also introduce substantial training challenges. As reported in RayZer (Tab.~7), training with explicit 3D leads to non-convergence. To address this, we propose a \textbf{fine-grained learning curriculum} built on the concept of \textit{visual overlap} between input views. We begin with samples of \textit{high visual overlap}, allowing the pose estimator to initialize from near-identity poses, and gradually reduce overlap to promote general 3D understanding.
When \textit{scaling} to heterogeneous training sources, visual overlap provides a natural and unified metric to adaptively align varying camera motion distributions, improving data consistency. Notably, we approximate visual overlap in an \textit{unsupervised} manner, keeping the framework entirely free from 3D annotations.

We systematically study \ours{}'s performance across different training data scales. We highlight key conclusions and summarize our contributions as follows:

\hspace*{0.05in}\textbullet~\ours{} is the first \textbf{truly self-supervised feedforward 3D Gaussian splatting reconstruction} model, trained \textbf{from scratch} with zero 3D annotation.

\hspace*{0.05in}\textbullet~\ours{} \textbf{outperforms prior visual representation learners}, \eg, DINOv3~\cite{simeoni2025dinov3}, CroCo v2~\cite{weinzaepfel2023croco}, VideoMAE V2~\cite{wang2023videomae}, and Perception Encoder~\cite{bolya2025perception}, on downstream 3D tasks (Tab.~\ref{tab:comparison_pose_depth}-\ref{tab:comparison_flow}), establishing \ours{} as a strong paradigm for \textit{spatial visual pre-training}.

\hspace*{0.05in}\textbullet~Compared with previous \underline{self-supervised} 3D vision models, \ours{} exhibits \textbf{stronger 3D understanding}, as evidenced by significantly improved unsupervised camera pose estimation (Tab.~\ref{tab:comparison_pose_and_nvs}) and the fine-tuning results on downstream 3D tasks (Tab.~\ref{tab:comparison_pose_depth}).

\hspace*{0.05in}\textbullet~Compared with state-of-the-art \underline{supervised} models, \eg, VGGT~\cite{wang2025vggt} (reproduced with matched architecture and training setups), \ours{} achieves \textbf{comparable or sometimes superior performance} (Tab.~\ref{tab:comparison_with_vggt}) and exhibits \textit{similar scaling behavior} (Tab.~\ref{tab:data_mixing_ablation}), despite being purely self-supervised.
\section{Related Work}

\textbf{Supervised Pose Estimation and 3D Reconstruction.} Early learning-based methods estimated relative camera poses from image pairs~\cite{balntas2018relocnet,banani2020novel,cai2021extreme,rockwell20228}, while later approaches extended this to multi-view reasoning across multiple inputs~\cite{zhang2022relpose,jiang2022few,jiang2024leap,lin2023relpose++,sinha2023sparsepose,wang2023posediffusion,zhang2024cameras}. Given posed images, 3D representations can be reconstructed either by direct regression~\cite{yu2021pixelnerf,jiang2022few,zhang2024gs} or by optimization-based mode seeking with diffusion models~\cite{zhou2023sparsefusion,zhao2024sparse}. More recently, several methods have unified pose estimation and 3D reconstruction by predicting pixel-aligned pointmaps~\cite{wang2023DUSt3R,duisterhof2024mast3r,wang2025continuous,wang2025vggt,zhao2025diffusionsfm}; these exhibit strong robustness under sparse inputs and generalize well across diverse domains~\cite{vuong2025aerialmegadepth}. Nevertheless, training such supervised models still relies on camera pose and dense depth annotations, typically obtained from traditional SfM systems (\eg, COLMAP~\cite{schonberger2016structure}), which can be inaccurate and ultimately limit performance.

Recent work has also investigated predicting 3D Gaussians~\cite{kerbl20233d} using photometric losses as (part of the) supervision. However, these methods remain \textit{de facto} supervised by 3D annotations, as they rely on ground-truth intrinsics~\cite{hongpf3plat,ye2024no,kang2025selfsplat} and/or target-view camera poses during training~\cite{smart2024splatt3r,ye2024no,kang2025selfsplat}, or require initialization and/or regularization from 3D-supervised models~\cite{smart2024splatt3r, jiang2025anysplat, huang2025no}. In contrast, \ours{} requires no 3D supervision and can be trained from scratch, making it \textit{truly self-supervised} -- while achieving even stronger performance.

\noindent\textbf{Self-supervised Novel-view Synthesis.} To alleviate the dependence on 3D supervision, another line of research learns scene representations directly from 2D images via novel-view synthesis. Early work predicted scene features from a single viewpoint and rendered target views as supervision~\cite{zhou2017unsupervised, wiles2020synsin,lai2021video, fu2023mononerf}. More recently, RUST~\cite{sajjadi2023rust}, RayZer~\cite{jiang2025rayzer}, and others~\cite{wang2025less, wang2025recollection, mitchel2025true} adopt learning-based latent rendering from multi-view inputs. However, these methods exhibit limited 3D awareness. For instance, RayZer learns view interpolation within an uninterpretable pose space. We build on RayZer but differ in three key respects: we adopt an explicit 3D representation (\ie, 3D Gaussians~\cite{kerbl20233d}), a more principled learning curriculum, and larger-scale training. We show that explicit 3D modeling yields more geometrically grounded representations, establishing a promising pre-training framework for downstream tasks that require 3D understanding.

Closely relevant to our work, DBARF~\cite{chen2023dbarf} explored a NeRF-based~\cite{mildenhall2021nerf} framework for self-supervised pose estimation and novel-view synthesis using a multi-stage pipeline. \ours{} offers a more streamlined and scalable framework, and goes beyond novel-view synthesis by investigating the learned representations for downstream tasks.

\noindent \textbf{Visual Pre-training for Representation Learning.} Prior work has made substantial progress in learning global image semantics through image-language association~\cite{radford2021learning, tschannen2023image, alayrac2022flamingo}, self-distillation~\cite{caron2021emerging, oquab2023dinov2, simeoni2025dinov3}, contrastive learning~\cite{he2020momentum, chen2020simple}, masked image modeling~\cite{pathak2016context, he2022masked}, and video-level temporal self-supervision~\cite{tong2022videomae, bardes2024revisiting, feichtenhofer2022masked}. However, learning \textit{3D-aware and geometrically grounded representations} remains underexplored, despite its potential to benefit 3D-related tasks where supervision is scarce. Recent efforts explore learning 3D awareness through proxy tasks such as latent-space novel-view synthesis~\cite{jiang2025rayzer, weinzaepfel2022croco, weinzaepfel2023croco}, but whether they enforce true 3D understanding remains unclear. In this work, \ours{} addresses this gap with explicit 3D modeling and a learning curriculum that enables effective scaling, yielding 3D-grounded representations.
\section{Approach}
\label{sec:approach}

\begin{figure*}[t]
  \centering
  \includegraphics[width=\textwidth]{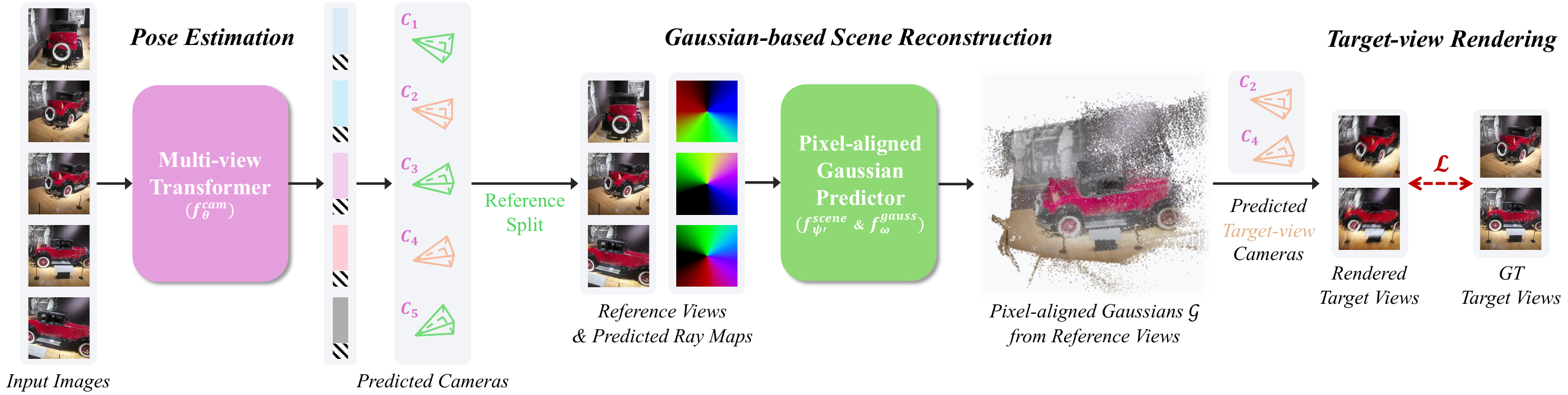}
  \vspace{-0.25in}
  \caption{\textbf{\ours{} Model \& Training.} \ours\ first predicts camera poses and intrinsics for all images. Then it follows RayZer~\cite{jiang2025rayzer} to split images into two sets. \ours{} predicts explicit 3D Gaussians as scene representation from the reference views ($\mathcal{I}_{\text{ref}}$), and renders the scene using self-predicted target-view ($\mathcal{I}_{\text{tgt}}$) cameras. Finally, \ours{} is trained with self-supervised photometric losses on target views.}
  \vspace{-0.1in}
  \label{fig:method}
\end{figure*}

From unlabeled multi-view images, \ours{} learns to predict camera parameters (poses and intrinsics) and \textbf{explicit} 3D scene geometry under self-supervision. Its learned representations can be further leveraged for downstream tasks, demonstrating \ours{}'s potential as a 3D-aware visual pre-training framework.

In the following, we first revisit RayZer~\cite{jiang2025rayzer}, the \textit{implicit} predecessor, and discuss its limitations (Sec.~\ref{sec:preliminaries}). Building on RayZer's core design while addressing these issues by leveraging \textit{Explicit} 3D modeling, we introduce \textit{E}-RayZer (Sec.~\ref{sec:architecture}). Finally, we present a sequence-level curriculum learning strategy based on visual overlap between frames to improve performance and scalability (Sec.~\ref{sec:curriculum}).

\subsection{Preliminaries: RayZer with Implicit 3D}
\label{sec:preliminaries}

RayZer splits all input images into two \textit{non-overlapping} subsets: an ``observed'' reference set ($\mathcal{I}_{\text{ref}}$) for latent scene inference, and a ``hidden'' target set ($\mathcal{I}_{\text{tgt}}$) for providing self-supervision. RayZer uses predicted cameras of target views ($\mathcal{I}_{\text{tgt}}$) to render the scene predicted from the reference views ($\mathcal{I}_{\text{ref}}$), and applies the photometric loss as self-supervision:
\begin{equation}
    \mathcal{L} = \Sigma_{(I, \hat{I}) \in ( \mathcal{I}_{\text{tgt}}, \hat{\mathcal{I}}_{\text{tgt}})} 
    \big(\texttt{MSE}(I, \hat{I}) + \lambda \cdot \texttt{Percep}(I, \hat{I}) \big), 
\label{eq:loss_render}
\end{equation}
where $\texttt{Percep}$ denotes the perceptual loss~\cite{johnson2016perceptual}. 

RayZer leverages transformers for pose estimation, latent (implicit) scene reconstruction, and rendering. It first predicts camera intrinsics and extrinsics for all input images $\mathcal{I} \in \mathbb{R}^{V \times H \times W \times 3}$ using a multi-view transformer $f_{\boldsymbol{\theta}}^{\text{cam}}$, as:
\begin{equation}
    (\mathbf{K},\, \mathbf{T}) = f_{\boldsymbol{\theta}}^{\text{cam}}(\mathcal{I}), \quad 
    \mathbf{T}_i = [\mathbf{R}_i \,|\, \mathbf{t}_i] \in SE(3),
\label{eq:pose_estimation}
\end{equation}
where $\mathbf{K} \in \mathbb{R}^{3 \times 3}$ is the intrinsics shared by all views, $\mathbf{T} \in \mathbb{R}^{V \times 4 \times 4}$ denotes the extrinsics, and $i = 1, \dots, V$ indexes the input images. 
Each camera $(\mathbf{K}, \mathbf{T}_i)$ is then converted into a pixel-aligned Plücker ray map $\mathbf{R}_i^{\text{plk}}$~\cite{plucker1865xvii,zhang2024cameras}.

To infer latent scene representations, 
RayZer tokenizes the concatenation (along the feature dimension) of image and rays for $\mathcal{I}_{\text{ref}}$ and updates a set of learnable scene tokens $\mathbf{z}_0^{\text{scene}}$ through another transformer $f_{\boldsymbol{\psi}}^{\text{scene}}$, as:
\begin{equation}
    \mathbf{z}^{\text{scene}}_{\text{ref}} = f_{\boldsymbol{\psi}}^{\text{scene}}
    \big(\mathbf{z}_0^{\text{scene}},\, \mathrm{Linear}(\mathcal{I}_{\text{ref}},\, \mathbf{R}_{\text{ref}}^{\text{plk}})\big),
\label{eq:scene_representation}
\end{equation}
where $\mathrm{Linear}(\cdot)$ denotes a patch-wise linear projection for fusing and tokenizing RGB and ray information. The resulting $\mathbf{z}^{\text{scene}}_{\text{ref}}$ represents the latent scene features.

For rendering, the self-predicted target-view Plücker ray maps are likewise tokenized and concatenated with the scene representation $\mathbf{z}^{\text{scene}}_{\text{ref}}$ (along the token dimension). These target-view ray tokens are refined via transformer $f_{\boldsymbol{\phi}}^{\text{rend}}$ and finally decoded to RGB images, as:
\begin{equation}
    \hat{\mathcal{I}}_{\text{tgt}} 
    = f_{\boldsymbol{\phi}}^{\text{rend}}
    \big(\mathbf{z}^{\text{scene}}_{\text{ref}},\, \mathrm{Linear}(\mathbf{R}_{\text{tgt}}^{\text{plk}})\big).
\label{eq:implicit_rendering}
\end{equation}
Then RayZer applies photometric self-supervision (Eq.~\ref{eq:loss_render}).

\noindent \textbf{Limitations of RayZer's Implicit 3D.}
RayZer achieves high-fidelity novel-view synthesis. However, \textbf{RayZer is not fully 3D-grounded}. Since its camera estimation ($f_{\boldsymbol{\theta}}^{\text{cam}}$), latent scene reconstruction ($f_{\boldsymbol{\psi}}^{\text{scene}}$), and rendering ($f_{\boldsymbol{\phi}}^{\text{rend}}$) modules are jointly learned from scratch, they only need to remain \textit{mutually compatible}, but are not guaranteed to be physically or spatially meaningful. This issue is further amplified by RayZer’s pure transformer-based architecture, which contains almost \textit{no 3D inductive bias} and thus possesses excessive flexibility to learn undesirable shortcut solutions. As evidenced by its imperfect camera pose distribution, RayZer relies on a mixture of true 3D understanding and video-interpolation priors to achieve high-quality synthesis. While this design suffices for novel-view synthesis, it limits RayZer’s potential as a \textit{spatial pre-training} framework for learning genuinely 3D-aware representations.

\subsection{\ours: Explicit 3D with Self-supervision}
\label{sec:architecture}

\noindent \textbf{Our Insights.} We argue that \textbf{3D inductive biases remain essential} for 3D representation learning, but must be introduced in ways that preserve \textbf{learning scalability}. 

We therefore inject a \textit{lightweight} 3D inductive bias through model design while keeping training fully self-supervised, striking a balance between \textit{3D awareness} and \textit{scalability}. Specifically, \ours{} replaces RayZer's latent scene representation with \textit{explicit} 3D geometry (\ie, 3D Gaussians~\cite{kerbl20233d}), providing geometric regularization that yields more grounded pose estimation, scene reconstruction, and latent representations.

\noindent \textbf{Overview.} As shown in Fig.~\ref{fig:method}, \ours\ first predicts the camera parameters for all images, and then infers pixel-aligned 3D Gaussians $\mathcal{G}$ from the reference view subset ($\mathcal{I}_{\text{ref}}$). Then \ours\ predicts the target view subset ($\mathcal{I}_{\text{tgt}}$), by rendering the 3D Gaussians predicted from $\mathcal{I}_{\text{ref}}$ under self-predicted cameras of $\mathcal{I}_{\text{tgt}}$.
Since 3D Gaussians support closed-form differentiable rendering, the latent rendering decoder used in RayZer (\ie, $f_{\boldsymbol{\phi}}^{\text{rend}}$ in Eq.~\ref{eq:implicit_rendering}) is no longer required. We now describe our \textit{key differences} from RayZer while elaborating on details.

\noindent \textbf{Gaussian-based Scene Reconstruction.} 
\ours{} first predicts the cameras for all views in a similar way with RayZer (besides differences in model architecture that  will be detailed later). Then, \ours{} directly transforms the ``posed'' reference views to pixel-aligned 3D Gaussians. We first encode posed reference views into latent tokens:
\begin{equation}
    \mathbf{s}_{\text{ref}} = f_{\boldsymbol{\psi}'}^{\text{scene}}
    \big(\mathrm{Linear}(\mathcal{I}_{\text{ref}},\, \mathbf{R}_{\text{ref}}^{\text{plk}})\big)
\label{eq:scene_representation_2}
\end{equation}
where $\mathbf{s}_{\text{ref}} \in \mathbb{R}^{K_{\text{ref}}hw\times C}$ denotes the updated image tokens of reference views after multi-view aggregation. In detail, $K_{\text{ref}}$ is the number of views in $\mathcal{I}_{\text{ref}}$, $h=H/p$ and $w=W/p$ are token number along height and width dimensions using a patch size of $p$, and $C$ is channel dimension of the latent space. Note that the complexity of global attention in Eq.~\ref{eq:scene_representation_2} is $\mathcal{O}((K_{\text{ref}}hw)^2)$, while it is $\mathcal{O}((K_{\text{ref}}hw + n_{\mathbf{z}})^2)$ for RayZer (Eq.~\ref{eq:scene_representation}), where $n_{\mathbf{z}}$ is the size for RayZer's scene token set.

Then, we use a lightweight decoder to transform the updated image tokens $\mathbf{s}_{\text{ref}}$ into per-pixel 3D Gaussian parameters along each camera ray across all reference views, as:
\begin{equation}
\begin{aligned}
    \mathcal{G} &= f_{\boldsymbol{\omega}}^{\text{gauss}}(\mathbf{s}_{\text{ref}}), \quad \text{where} \\
    \mathcal{G} &= \big\{\, g_i = (d_i,\, \mathbf{q}_i,\, \mathbf{C}_i,\, \mathbf{s}_i,\, \alpha_i) \,\big\}_{i=1}^{K_{\text{ref}} \times H \times W}.
\end{aligned}
\label{eq:gaussian_prediction}
\end{equation}
These parameters include the distance along the ray $d_i \in \mathbb{R}$, orientation represented as a quaternion $\mathbf{q}_i \in \mathbb{R}^4$, spherical harmonic coefficients $\mathbf{C}_i \in \mathbb{R}^{(d_{\text{SH}} + 1)^2 \times 3}$, scale $\mathbf{s}_i \in \mathbb{R}^3$, and opacity $\alpha_i \in \mathbb{R}$. The predicted 3D Gaussians collectively represent the scene geometry.

We then use \ours{}'s self-predicted target views cameras, denoted as $\mathcal{C}_{\text{tgt}} = \{\, (\mathbf{K}, \mathbf{T}_i) \mid i \in \mathcal{I}_{\text{tgt}} \,\}$, to render the 3D Gaussians $\mathcal{G}$ and get prediction of target views, as:
\begin{equation}
   \hat{\mathcal{I}}_{\text{tgt}} = \pi(\mathcal{G}, \mathcal{C}_{\text{tgt}}), 
\end{equation}
where $\pi$ denotes the differentiable rendering operations of 3D Gaussians. Note that we modify gsplat~\cite{ye2025gsplat} to support gradient back-propagation to camera intrinsics $\textbf{K}$. Compared with RayZer, this design improves both \textit{rendering efficiency} and \textit{3D-awareness} by removing the need to learn a transformer-based renderer. Finally, we apply photometric loss on the rendered target views as Eq.~\ref{eq:loss_render}.

\noindent \textbf{Avoiding Undesirable View Interpolation.}
As discussed in Sec.~\ref{sec:preliminaries}, RayZer tends to learn undesirable frame interpolation cues as shortcut solutions. We identify a main cause as its use of \textit{image index embeddings} to associate image tokens with corresponding camera tokens for camera estimation, which provides a strong cue for learning interpolation.

In \ours{}, we remove the image index embeddings entirely. We adopt a VGGT-style~\cite{wang2025vggt} multi-view transformer with alternating local-global attention, where the local attention boundary naturally defines the association relationship.
Different from the original VGGT, \ours{} performs \textit{pairwise pose prediction}: camera tokens from a canonical view and a target view are concatenated to regress their relative camera pose. Consequently, \ours{} does not require different camera/register tokens for canonical and non-canonical views. This architectural design is applied to both the transformer used for camera estimation ($f_{\boldsymbol{\theta}}^{\text{cam}}$) and that for scene reconstruction ($f_{\boldsymbol{\psi}'}^{\text{scene}}$).

\begin{figure}[t]
  \centering
  \includegraphics[width=\columnwidth]{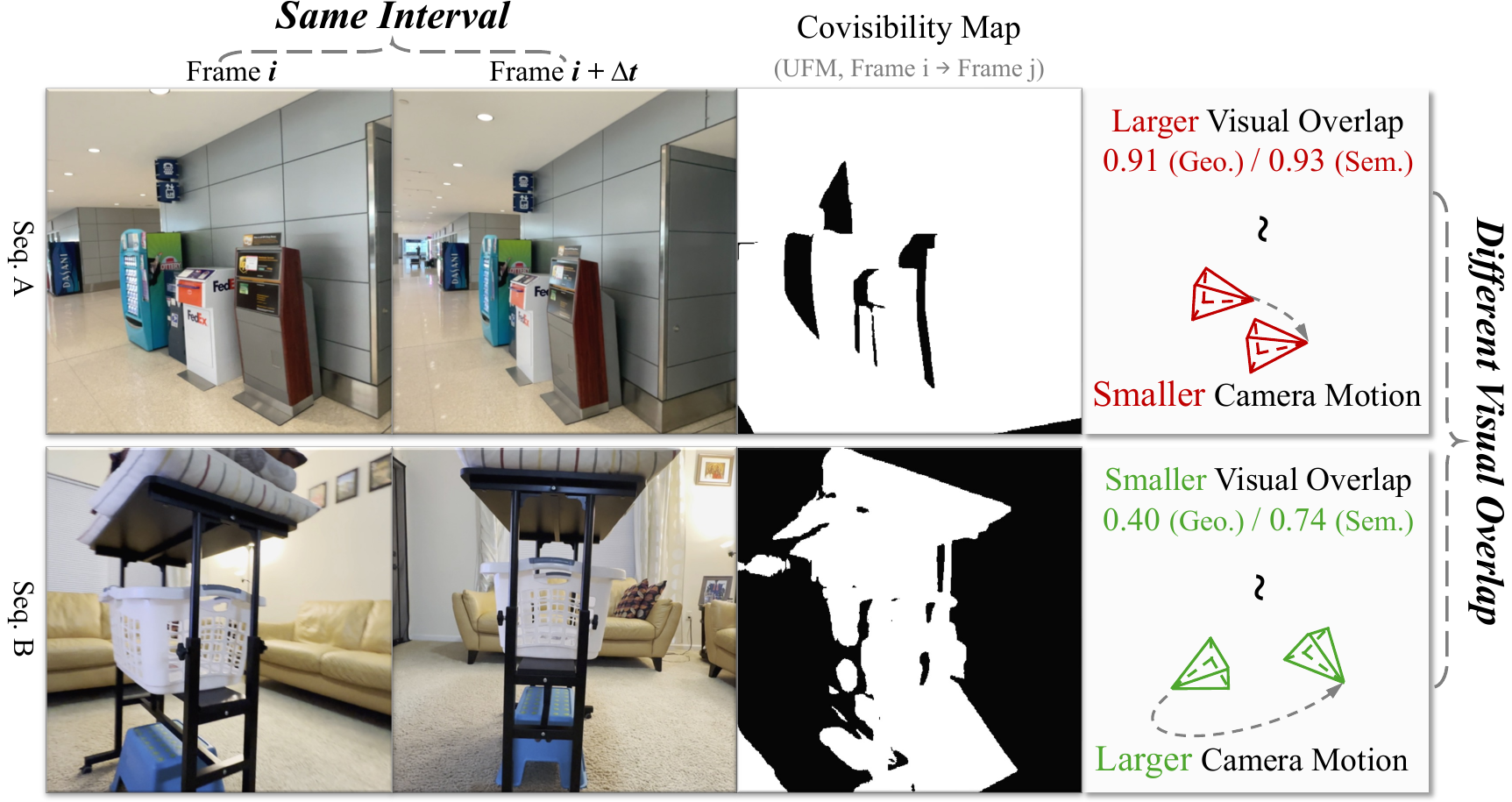}
  \vspace{-0.25in}
  \caption{
  \textbf{Different Visual Overlaps under the Same Frame Interval.} Two sequences from DL3DV~\cite{ling2024dl3dv} share the same frame interval yet exhibit drastically different levels of visual overlap. Our proposed semantic and geometric overlap metrics more accurately capture the true difficulty (or camera motion) of each sequence.}
  \vspace{-0.15in}
  \label{fig:curriculum}
\end{figure}

\subsection{Sequence Curriculum Based on Visual Overlap}
\label{sec:curriculum}

As \ours{} leverages explicit scene representation, it suffers from harder convergence when \textbf{trained from scratch}. To stabilize training, we propose a learning curriculum based on the concept of \textit{visual overlap} between input views, providing \textit{fine-grained control} over training data difficulty. This curriculum also adaptively aligns the data distributions across diverse data sources, making \ours{} more scalable to heterogeneous training resources.

We highlight that \ours{}’s learning curriculum fundamentally differs from that of RayZer, which is based on fixed frame-index intervals.
As illustrated in Fig.~\ref{fig:curriculum}, RayZer’s interval-based sampling provides only an inaccurate and inflexible approximation of visual overlap, is hard-coded and thus not scalable to heterogeneous resources.

We then describe the two key steps for constructing our learning curriculum: \textit{data labeling} and \textit{sampling}.
We introduce two variants of visual-overlap labeling tools: a \textit{geometric} version that computes actual covisibility, and a \textit{semantic} version as an unsupervised approximation of it.

\textbf{\emph{Labeling.}} For each training sequence $u$ (from any data resource), we compute a spacing profile by \textit{uniformly} sampling a small set of frame triplets for each spacing $\Delta t$, as $\mathcal{T}_u(\Delta t)=\{(i,\, i+\Delta t,\, i+2\Delta t)\}$, and averaging the two pairwise overlaps $o(\cdot,\cdot)$ per triplet:
\begin{equation}
o_{\text{tri}}(i,\Delta t) \;=\; \tfrac{1}{2}\Big( o(i,\, i+\Delta t) \;+\; o(i+\Delta t,\, i+2\Delta t) \Big).
\label{eq:triplet_overlap}
\end{equation}
Averaging $o_{\text{tri}}(i,\Delta t)$ over all sampled triplets yields the per-sequence profile $O_u(\Delta t)$, characterizing how overlap (and consequently difficulty) varies with frame index spacing.

\begin{figure*}[t]
  \centering
  \includegraphics[width=\textwidth]{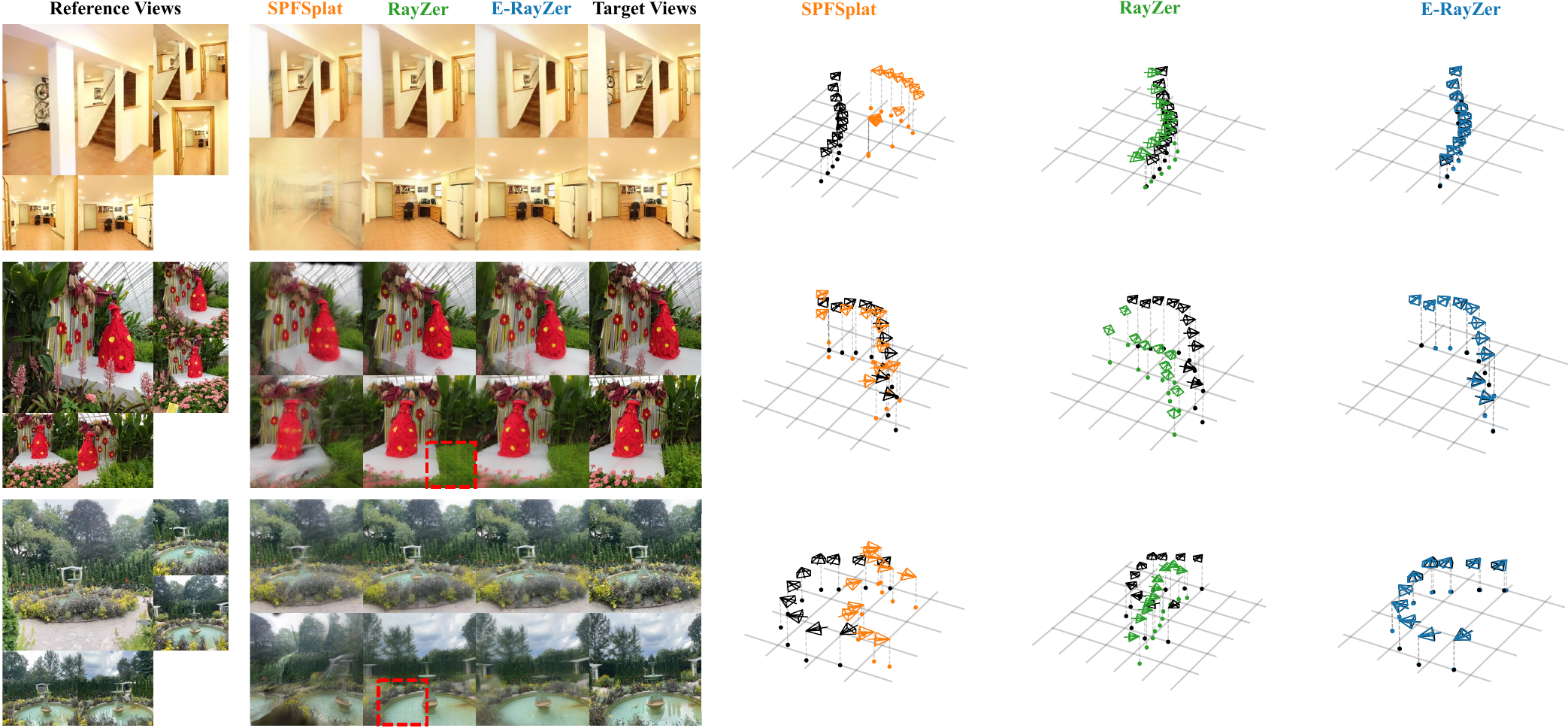}
  \vspace{-0.27in}
  \caption{\textbf{Visual Comparison with (Partially) Self-supervised Methods.} We include results on both novel-view synthesis (left) and pose estimation (right), where \ours{} outperforms baselines on pose accuracy, showing its grounded 3D understanding. \ours{} also outperforms RayZer on low-texture regions (highlighted w/ red box) on NVS, a case where RayZer's view interpolation cannot handle.}
  \label{fig:main_compare}
  \vspace{-0.1in}
\end{figure*}

\begin{table*}[t]
  \centering
  \caption{\textbf{Comparison with (Partially) Self-supervised Methods on Novel-view Synthesis (NVS) and Pose Estimation.} We report PSNR for NVS and RPA$_\uparrow$@5°/15°/30° for pose estimation. RayZer~\cite{jiang2025rayzer} and \ours{} are fully self-supervised methods trained from scratch, while SPFSplat~\cite{hongpf3plat} is initialized from MASt3R~\cite{duisterhof2024mast3r}, which itself is trained under dense 3D supervision on 14 datasets.}
  \vspace{-0.12in}
  \label{tab:comparison_pose_and_nvs}
  \resizebox{\textwidth}{!}{
  \begin{tabular}{lcccccccccccccc}
    \toprule
    \multirow{2}{*}{\textbf{Method}} & \multirow{2}{*}{\textbf{Self-supervised?}} & \multirow{2}{*}{\textbf{Training Data}}  &
    \multicolumn{4}{c}{\textbf{WildRGB-D}~\cite{xia2024rgbd}} &
    \multicolumn{4}{c}{\textbf{ScanNet++}~\cite{yeshwanth2023scannet++}} &
    \multicolumn{4}{c}{\textbf{DL3DV}~\cite{ling2024dl3dv}} \\
    \cmidrule(lr){4-7} \cmidrule(lr){8-11} \cmidrule(lr){12-15}
    & & & PSNR$_\uparrow$ & @5°$_\uparrow$ & @15°$_\uparrow$ & @30°$_\uparrow$
    & PSNR$_\uparrow$ & @5°$_\uparrow$ & @15°$_\uparrow$ & @30°$_\uparrow$
    & PSNR$_\uparrow$ & @5°$_\uparrow$ & @15°$_\uparrow$ & @30°$_\uparrow$ \\
    \midrule
    SPFSplat~\cite{huang2025no} & \cellcolor{tablered2}\ding{55} (MASt3R ini.) & RE10K~\cite{zhou2018stereo} (+ extra) &
    16.7 & 31.5 & 58.0 & 69.8 &
    14.0 & \textbf{2.5} & 11.8 & 30.3 &
    15.1 & 19.5 & 40.6 & 50.5 \\
    \ours{} (ours) & \cellcolor{tablegreen}\ding{51} & RE10K~\cite{zhou2018stereo} &
    \textbf{21.0} & \textbf{40.3} & \textbf{89.4} & \textbf{96.5} &
    \textbf{17.5} & 1.1  & \textbf{13.3} & \textbf{37.3} &
    \textbf{17.3} & \textbf{21.2} & \textbf{55.0} & \textbf{72.7} \\
    \midrule
    RayZer~\cite{jiang2025rayzer} & \cellcolor{tablegreen}\ding{51} & \multirow{2}{*}{DL3DV~\cite{ling2024dl3dv}} &
    \textbf{25.9} & 0.0  & 0.2  & 6.5  &
    \textbf{20.5} & 0.0  & 0.7  & 6.2  &
    \textbf{21.4} & 0.0  & 0.6  & 6.2  \\
    \ours\ (ours) & \cellcolor{tablegreen} \ding{51} & &
    24.3 & \textbf{84.5} & 
    \textbf{98.4} & \textbf{99.3} & 
    20.1 & \textbf{7.7}  & 
    \textbf{33.6} & \textbf{63.0} & 
    20.3 & \textbf{72.0} & 
    \textbf{88.4} & \textbf{93.5} \\ \midrule
    RayZer~\cite{jiang2025rayzer} & \cellcolor{tablegreen}\ding{51} &  \multirow{2}{*}{7 datasets}&
    \textbf{26.7} & 0.2  & 9.3  & 43.6 &
    \textbf{21.5} & 0.0  & 0.9  & 9.0  &
    \textbf{20.8} & 0.0  & 1.9  & 17.0 \\
    \ours\ (ours) & \cellcolor{tablegreen} \ding{51} & &
    24.9 & \textbf{90.8} & 
    \textbf{98.6} & \textbf{99.3} & 
    20.7 & \textbf{5.7}  & 
    \textbf{34.8} & \textbf{63.7} & 
    19.7 & \textbf{59.9} & 
    \textbf{82.9} & \textbf{90.2} \\
    \bottomrule
  \end{tabular}}
  \vspace{-0.15in}
\end{table*}

\begin{table*}[t]
  \centering
  \caption{\textbf{Comparison with Supervised VGGT~\cite{wang2025vggt} on Pose Estimation. \ours{}'s pre-training improves VGGT performance} (last row), forming an effective self-supervised pre-training and supervised post-training paradigm. We report pose accuracy RPA$_\uparrow@5^\circ$/$15^\circ$. Both models are trained on DL3DV~\cite{ling2024dl3dv} and evaluated on DL3DV \& eight out-of-domain datasets for zero-shot testing. Models are labeled as \textcolor[rgb]{0.35,0.8,0.35}{self-supervised} or \textcolor{tablered2!250}{supervised}. VGGT* denotes our re-implementation with \ours{}'s pairwise camera head. Results are color-ranked from red to yellow, and we \underline{underline} the results that our self-supervised \ours{} surpasses supervised VGGT*.} 
  \vspace{-0.12in}
  \label{tab:comparison_with_vggt}
  \resizebox{\textwidth}{!}{
  \renewcommand{\arraystretch}{1.02}
  \begin{tabular}{@{}p{3.6cm}@{}cc|cccccccccccccccc}
    \toprule
    \multirow{3}{*}{\textbf{Method}} & \multicolumn{2}{c|}{\textit{In-domain}} & \multicolumn{16}{c}{\textit{Out-of-domain (Zero-shot Generalization)}}
    \\
     &
    \multicolumn{2}{c|}{\textbf{DL3DV}~\cite{reizenstein2021common}} &
    \multicolumn{2}{c}{\textbf{RE10K}~\cite{zhou2018stereo}} &
    \multicolumn{2}{c}{\textbf{CO3Dv2}~\cite{reizenstein2021common}} &
    \multicolumn{2}{c}{\textbf{WildRGB-D}~\cite{xia2024rgbd}} &
    \multicolumn{2}{c}{\textbf{7-Scenes}~\cite{shotton2013scene}} &
    \multicolumn{2}{c}{\textbf{CamLand}~\cite{kendall2017geometric}} &
    \multicolumn{2}{c}{\textbf{BlendedMVS}~\cite{yao2020blendedmvs}} &
    \multicolumn{2}{c}{\textbf{NAVI}~\cite{jampani2024navi}} &
    \multicolumn{2}{c}{\textbf{ScanNet++}~\cite{yeshwanth2023scannet++}} \\
    \cmidrule(lr){2-3} \cmidrule(lr){4-5} \cmidrule(lr){6-7} \cmidrule(lr){8-9} \cmidrule(lr){10-11} \cmidrule(lr){12-13} \cmidrule(lr){14-15} \cmidrule(lr){16-17} \cmidrule(lr){18-19}
    & @5°$_\uparrow$ & @15°$_\uparrow$ & @5°$_\uparrow$ & @15°$_\uparrow$ & @5°$_\uparrow$ & @15°$_\uparrow$ & @5°$_\uparrow$ & @15°$_\uparrow$ & @5°$_\uparrow$ & @15°$_\uparrow$ & @5°$_\uparrow$ & @15°$_\uparrow$ & @5°$_\uparrow$ & @15°$_\uparrow$ & @5°$_\uparrow$ & @15°$_\uparrow$ & @5°$_\uparrow$ & @15°$_\uparrow$ \\
    \hline
    \solidcell{tablegreen}{\ours{} (ours)}
    & \cellcolor{yellow}72.0 & \cellcolor{yellow}88.4 & \cellcolor{orange}\underline{83.0} & \cellcolor{yellow}96.8 & \cellcolor{orange}\underline{19.1} & \cellcolor{yellow}61.8 & \cellcolor{orange}\underline{51.1} & \cellcolor{orange}\underline{82.3} & \cellcolor{orange}\underline{38.8} & \cellcolor{yellow}78.0 & \cellcolor{orange}\underline{18.1} & \cellcolor{orange}\underline{62.9} & \cellcolor{orange}\underline{22.9} & \cellcolor{orange}\underline{46.8} & \cellcolor{orange}\underline{20.7} & \cellcolor{orange}\underline{57.8} & \cellcolor{orange}\underline{7.7} & \cellcolor{yellow}33.6\\ 
    
    \solidcell{tablered2}{VGGT*}
    & \cellcolor{orange}79.6 & \cellcolor{orange}94.2 & \cellcolor{yellow}80.4 & \cellcolor{orange}97.9 & \cellcolor{yellow}16.0 & \cellcolor{orange}64.3 & \cellcolor{yellow}32.5 & \cellcolor{yellow}76.2 & \cellcolor{yellow}34.7 & \cellcolor{tablered}\textbf{83.6} & \cellcolor{yellow}11.1 & \cellcolor{yellow}49.8 & \cellcolor{yellow}17.0 & \cellcolor{yellow}42.8 & \cellcolor{yellow}14.3 & \cellcolor{yellow}54.5 & \cellcolor{yellow}6.7 & \cellcolor{orange}39.8 \\
    \midrule
    \tikz[baseline=(n.base)]{
      \node[
        inner sep=0.4ex, outer sep=0pt,
        minimum width=\linewidth,
        minimum height=1.6em,
        shade,
        left color=tablegreen!110,
        right color=tablered2!110
      ] (n) {\makebox[\linewidth][l]{\strut \ours{}→VGGT*}};
    }
    & \cellcolor{tablered}\textbf{87.3} & \cellcolor{tablered}\textbf{96.6} & 
    \cellcolor{tablered}\textbf{85.3} & \cellcolor{tablered}\textbf{98.4} & 
    \cellcolor{tablered}\textbf{25.3} & \cellcolor{tablered}\textbf{72.2} & 
    \cellcolor{tablered}\textbf{56.2} & \cellcolor{tablered}\textbf{91.4} & 
    \cellcolor{tablered}\textbf{43.8} & \cellcolor{orange}82.8 & 
    \cellcolor{tablered}\textbf{30.2} & \cellcolor{tablered}\textbf{75.6} & 
    \cellcolor{tablered}\textbf{29.2} & \cellcolor{tablered}\textbf{52.2} & 
    \cellcolor{tablered}\textbf{26.9} & \cellcolor{tablered}\textbf{64.3} & 
    \cellcolor{tablered}\textbf{14.3} & \cellcolor{tablered}\textbf{53.8}\\
    \bottomrule
  \end{tabular}}
  \vspace{-0.14in}
\end{table*}

\textbf{\emph{Training-time Sampling.}} Given curriculum progress $s\!\in\![0,1]$, we use a visual overlap lower limit of
$o(s) \;=\; s\,o_{\min} \;+\; (1-s)\,o_{\max}$,
so that it decreases over training. We then obtain the sequence-specific spacing $\Delta t_u(s)$ by \textit{looking up} the precomputed table $\{(\Delta t_k,\, O_u(\Delta t_k))\}$ and \textit{linearly interpolating} between the nearest entries. Finally, the temporal span of the sampled sequence follows $t=(K_{\text{ref}}-1)\,\Delta t_u(s)$.

\textbf{\emph{Instantiations.}} We instantiate $o$ with two alternatives -- \textit{geometric overlap} (UFM~\cite{zhang2025ufm} covisibility, which is trained with 3D annotations) and \textit{semantic overlap} (DINOv2~\cite{oquab2023dinov2} cosine similarity, which is trained with self-supervision):
\begin{equation}
\begin{aligned}
o_{\text{sem}}(i,j) &= \cos\!\big(\phi_{\text{DINO}}(I_i),\, \phi_{\text{DINO}}(I_j)\big),\\
o_{\text{geo}}(i,j) &= \mathrm{Cov}_{\text{UFM}}(I_i,\, I_j).
\end{aligned}
\end{equation}

In Sec.~\ref{sec:exp_ablation}, we show that both the semantic and geometric  curricula outperform RayZer’s interval-based curriculum, and that the two variants perform comparably.

\section{Experiments}
\label{sec:experiments}

\begin{table}[t]
\centering
\small
\caption{\textbf{Probing 3D Spatial Awareness of Learned Representations on Multi-view Depth and Pose Estimation.}  
We evaluate the learned representations via both frozen-backbone and fully supervised finetuning on ScanNet++~\cite{yeshwanth2023scannet++} and BlendedMVS~\cite{yao2020blendedmvs}, which are not included in pre-training for any model. The best results are shown in \textbf{bold}, and the second-best are \underline{underlined}. The experiments only use the \textit{encoders} of RayZer~\cite{jiang2025rayzer} and \ours{}.}
\vspace{-0.1in}
\label{tab:comparison_pose_depth}
\footnotesize
\renewcommand{\arraystretch}{0.98}
\resizebox{\linewidth}{!}{
\begin{tabular}{c l l cc|cc}
\toprule
& & \multirow{2}{*}{\textbf{Method}} & \multicolumn{2}{c}{\textit{Depth}} & \multicolumn{2}{c}{\textit{Camera Pose}} \\
\cmidrule(lr){4-5} \cmidrule(lr){6-7}
& &  & \textbf{AbsRel}$_\downarrow$
& \textbf{$\boldsymbol{\delta}{<}$1.25}$_\uparrow$
& \textbf{RPA@5°}$_\uparrow$ & \textbf{RPA@15°}$_\uparrow$ \\
\midrule
\parbox[t]{2mm}{\multirow{14}{*}{\rotatebox[origin=c]{90}{\footnotesize \textbf{ScanNet++}~\cite{yeshwanth2023scannet++}}}}
& \multirow{7}{*}{\rotatebox[origin=c]{90}{\footnotesize Frozen}}
&   DINOv2~\cite{oquab2023dinov2}              & 0.193 & 74.9 & 0.8 & 9.5  \\
& & DINOv3~\cite{simeoni2025dinov3} & 0.201 & 73.2 & 0.4 & 10.0 \\
& & Percep. Encoder~\cite{bolya2025perception} & 0.203 & 73.2 & 0.5 & 8.5  \\
& & CroCo v2~\cite{weinzaepfel2023croco}       & 0.203 & 73.0 & 1.4 & 15.1 \\
& & VideoMAE V2~\cite{wang2023videomae}        & 0.175 & 76.3 & 0.1 & 6.6  \\
& & RayZer~\cite{jiang2025rayzer}              & \underline{0.161} & \underline{79.3} & \underline{4.7} & \underline{27.4}  \\
&  & \ours{} (ours)                            & \textbf{0.116} & \textbf{87.1} & \textbf{13.8} & \textbf{49.5} \\
\cmidrule(lr){2-7}
& \multirow{7}{*}{\rotatebox[origin=c]{90}{\footnotesize Full-finetune}}
&   DINOv2~\cite{oquab2023dinov2}              & 0.178 & 78.2 & 3.3  & 19.6 \\
& & DINOv3~\cite{simeoni2025dinov3} & 0.176 & 78.7 &  4.0 & 22.3 \\
& & Percep. Encoder~\cite{bolya2025perception} & 0.181 & 77.8 & 2.9  & 20.0 \\
& & CroCo v2~\cite{weinzaepfel2023croco}       & 0.177 & 78.2 & 3.8  & 20.8 \\
& & VideoMAE V2~\cite{wang2023videomae}        & \underline{0.076} & \underline{93.9} & 12.8 & 51.4 \\
& & RayZer~\cite{jiang2025rayzer}              & 0.077 & \underline{93.9} & \underline{21.5} & \underline{60.6} \\
& & \ours{} (ours)                             & \textbf{0.059} & \textbf{95.1} & \textbf{22.7} & \textbf{64.3} \\
\midrule \midrule
\parbox[t]{2mm}{\multirow{14}{*}{\rotatebox[origin=c]{90}{\footnotesize \textbf{BlendedMVS}~\cite{yao2020blendedmvs}}}}
& \multirow{7}{*}{\rotatebox[origin=c]{90}{\footnotesize Frozen}}
&   DINOv2~\cite{oquab2023dinov2}              & 0.366 & 50.5 & 1.1  & 8.0  \\
& & DINOv3~\cite{simeoni2025dinov3} & 0.397 & 49.1 & 1.2 & 6.8\\
& & Percep. Encoder~\cite{bolya2025perception} & 0.385 & 49.9  & 1.2 & 6.2 \\
& & CroCo v2~\cite{weinzaepfel2023croco}       & 0.412 & 47.7 & 1.6  & 12.6 \\
& & VideoMAE V2~\cite{wang2023videomae}        & 0.371 & 49.4 & 1.0  & 6.2  \\
& & RayZer~\cite{jiang2025rayzer}              & \underline{0.351} & \underline{52.6} & \underline{16.7} & \underline{34.5} \\
& & \ours{} (ours)                               & \textbf{0.245} & \textbf{68.3} & \textbf{26.5} & \textbf{45.8} \\
\cmidrule(lr){2-7}
& \multirow{7}{*}{\rotatebox[origin=c]{90}{\footnotesize Full-finetune}}
&   DINOv2~\cite{oquab2023dinov2}             & 0.353 & 52.5 & 1.8  & 12.8 \\
& & DINOv3~\cite{simeoni2025dinov3} & 0.349 & 52.1 & 1.7 & 15.3\\
& & Percp. Encoder~\cite{bolya2025perception} & 0.370 & 50.3 & 2.1 & 11.6 \\
& & CroCo v2~\cite{weinzaepfel2023croco}      & 0.369 & 51.2 & 2.8  & 15.9 \\
& & VideoMAE V2~\cite{wang2023videomae}       & 0.197 & 75.9 & 17.3 & 45.5 \\
& & RayZer~\cite{jiang2025rayzer}             & \underline{0.194} & \underline{77.7} & \underline{26.1} & \underline{50.2} \\
& & \ours{} (ours)                            & \textbf{0.148} & \textbf{82.8} & \textbf{36.2} & \textbf{58.8} \\
\bottomrule
\end{tabular}
}
\vspace{-0.1in}
\end{table}

We first describe the experimental setups in Sec.~\ref{sec:exp_setup}. We then evaluate \ours{} in two aspects: as a self-supervised model for pose estimation and 3D reconstruction (Sec.~\ref{sec:exp_pose_and_nvs}), and as a spatial visual pre-training framework for downstream tasks (Sec.~\ref{sec:exp_pretraining}). Finally, we ablate the key design choices of \ours{} (Sec.~\ref{sec:exp_ablation}).

\subsection{Experimental Setup}
\label{sec:exp_setup}

\noindent \textbf{Implementation Details.} \ours{} is trained with 10 input images, where 5 are used as reference views and 5 as target views. 
During training, we follow a linear decay in visual-overlap scores: $1.0 \rightarrow 0.5$ for geometric-overlap scheduling and $1.0 \rightarrow 0.75$ for semantic-overlap scheduling. 
For a fair comparison, we align RayZer with \ours{} using the better model architecture and the novel training curriculum.
For other baselines, we use official checkpoints and provide specific implementation details in the corresponding subsections. See more details in the supplementary material.

\noindent \textbf{Metrics.} For pose estimation, we report relative pose accuracy (RPA) at thresholds of 5$^\circ$, 15$^\circ$, and 30$^\circ$, which jointly reflects rotation and translation accuracy. For novel-view synthesis, we use standard PSNR. For depth estimation, we evaluate absolute relative error (AbsRel) and $\delta < 1.25$, following Depth Anything~\cite{yang2024depthanything}. For pairwise flow prediction, we report the average end-point error (EPE) and the proportion of outlier flow predictions under thresholds of 1px, 2px, and 5px, following UFM~\cite{zhang2025ufm}.

\noindent \textbf{Datasets.} \emph{Training.} We present results of \ours{} trained on both single-dataset and multi-dataset settings. The single-dataset variants are trained exclusively on RealEstate10K~\cite{sargent2023zeronvs} or DL3DV~\cite{ling2024dl3dv}, while the multi-dataset variant is trained on a mixture of seven datasets: DL3DV~\cite{ling2024dl3dv}, CO3Dv2~\cite{reizenstein2021common}, RealEstate10K~\cite{zhou2018stereo}, MVImgNet~\cite{yu2023mvimgnet}, ARKitScenes~\cite{baruch2021arkitscenes}, WildRGB-D~\cite{xia2024rgbd}, and ACID~\cite{liu2021infinite}, covering diverse indoor and outdoor sequences.

\emph{Evaluation.} We primarily evaluate pose estimation and novel-view synthesis on WildRGB-D, DL3DV test set, and the out-of-distribution (OOD) ScanNet++~\cite{yeshwanth2023scannet++}. To assess the generalization of the learned representations (Sec.~\ref{sec:exp_pretraining}), we evaluate on OOD ScanNet++ and BlendedMVS~\cite{yao2020blendedmvs} for pose and depth estimation, and StaticThings3D~\cite{schroppel2022benchmark} for pairwise flow prediction.

\vspace{-0.02in}
\subsection{Pose Estimation and Novel-view Synthesis}
\label{sec:exp_pose_and_nvs}
\vspace{-0.02in}

\noindent \textbf{Baselines and Setups.} We compare against SPFSplat~\cite{huang2025no} and RayZer~\cite{jiang2025rayzer}. Notably, SPFSplat is initialized from the supervised MASt3R~\cite{leroy2024grounding} model, and thus is not truly self-supervised; while \ours{} and RayZer are trained from scratch under self-supervision. We evaluate pose accuracy on all images and assess novel-view synthesis quality on the target views rendered with predicted camera poses.

\noindent \textbf{Results.} As shown in Tab.~\ref{tab:comparison_pose_and_nvs}, \ours{} consistently outperforms SPFSplat~\cite{huang2025no} on most metrics, despite being truly self-supervised. Moreover, \ours{} significantly surpasses RayZer~\cite{jiang2025rayzer} in pose estimation across all setups while achieving comparable novel-view synthesis quality. These results suggest that the \textit{explicit} 3D modeling in \ours{} yields more geometrically meaningful pose representations, whereas RayZer's \textit{implicit} approach over-optimizes for view synthesis quality without being truly 3D-aware, resulting in a less interpretable pose space. Qualitative comparisons in Fig.~\ref{fig:main_compare} further support these findings.

\begin{table}[t]
\centering
\caption{\textbf{Probing 2.5D Spatial Awareness of Learned Representations on Pairwise Flow Estimation.} We evaluate on StaticThings3D~\cite{schroppel2022benchmark}, an out-of-distribution synthetic dataset. All models are fully finetuned under flow supervision. The best results are shown in \textbf{bold}, and the second-best are \underline{underlined}.}
\vspace{-0.1in}
\label{tab:comparison_flow}
\scriptsize
\setlength{\heavyrulewidth}{0.08em}
\setlength{\lightrulewidth}{0.06em}
\setlength{\cmidrulewidth}{0.06em}
\renewcommand{\arraystretch}{0.98}
\begin{tabular*}{\linewidth}{@{\extracolsep{\fill}}lcccc}
\toprule
\multirow{2}{*}{\textbf{Method}} & \textit{Error} & \multicolumn{3}{c}{\textit{Outlier Ratio}}\\
\cmidrule(lr){2-2} \cmidrule(lr){3-5}
& \textbf{EPE}$_\downarrow$ & @\textbf{1px}$_\downarrow$ & @\textbf{2px}$_\downarrow$ & @\textbf{5px}$_\downarrow$ \\
\midrule
CroCo v2~\cite{weinzaepfel2023croco} & 1.273 & 17.7 & 8.7  & 3.8 \\
VideoMAE V2~\cite{wang2023videomae}  & 2.028 & 42.7 & 22.1 & 6.9 \\
RayZer~\cite{jiang2025rayzer}        & \textbf{1.105} & \textbf{13.4} & \textbf{6.6} & \textbf{2.8} \\
\ours{} (ours) & \underline{1.254}   & \underline{16.9} & \underline{7.8} & \underline{3.1} \\
\bottomrule
\end{tabular*}
\vspace{-0.2in}
\end{table}

\vspace{-0.04in}
\subsection{\ours{} as Self-supervised Pre-training}
\label{sec:exp_pretraining}
\vspace{-0.01in}

We validate \ours{} as a self-supervised spatial visual pre-training framework.
First, we show that its performance is comparable to the supervised VGGT and that \ours{} pre-training further enhances VGGT (Sec.~\ref{sec: exp_ours_and_vggt}).
We then probe the learned features on downstream tasks to verify \ours{}’s representation quality (Sec.~\ref{sec:exp_downstream}).

\vspace{-0.01in}
\subsubsection{\ours{} Initialization Benefits Supervised Model}
\label{sec: exp_ours_and_vggt}
\vspace{-0.01in}

\noindent \textbf{Baselines and Setups.} We compare with the state-of-the-art supervised model VGGT~\cite{wang2025vggt}. Note that we train it using the same data and architecture with \ours{} for an apple-to-apple comparison, denoted as VGGT*.

\noindent \textbf{\ours{} is Comparable with Supervised VGGT*.} First two rows of Tab.~\ref{tab:comparison_with_vggt} show that \ours{} outperforms  VGGT* on several out-of-domain datasets (\eg, WildRGB-D~\cite{xia2024rgbd}, CamLand~\cite{kendall2017geometric}, and BlendedMVS~\cite{yao2020blendedmvs}). Moreover, \ours{} almost consistently achieves higher accuracy on RPA@5$^\circ$, a stricter metric, suggesting better precision in pose prediction. The results demonstrate the strong performance of \ours{} as a self-supervised method without using any 3D annotations for training.

\noindent \textbf{Effectiveness of Pre-training.} As shown in last two rows of Tab.~\ref{tab:comparison_with_vggt}, initializing VGGT* with \ours{} weights yields significant improvements over training from scratch, confirming that \ours{} serves as an effective pre-training framework for visual geometry learning. The results also suggest that the learned knowledge of our self-supervised and supervised methods are highly complementary (they are trained on same data but pre-training still helps), showing the great potential of spatial visual pre-training.

\begin{table*}[t]
  \centering
  \small
  \caption{\textbf{Ablation on Data Mixing and Scaling.} 
  We compare our E-RayZer with supervised VGGT*~\cite{wang2025vggt} on varying training data settings. 
  We color-rank the results from red to yellow \textbf{for each model itself} across training data, thus \textbf{the color distribution reflects their scaling behavior}. We also \underline{underline} the results where \textcolor[rgb]{0.35,0.8,0.35}{self-supervised} \ours{} outperforms \textcolor{tablered2!250}{supervised} VGGT* (for each training data).}
  \vspace{-0.12in}
  \label{tab:data_mixing_ablation}
  \renewcommand{\arraystretch}{0.98}
  \resizebox{\textwidth}{!}{
  \begin{tabular}{llcccccccccccccccc}
    \toprule
    \multirow{2}{*}{\textbf{Training Data}} &
    \multirow{2}{*}{\textbf{Method}} &
    \multicolumn{4}{c}{\textbf{NAVI}~\cite{jampani2024navi}} &
    \multicolumn{4}{c}{\textbf{CO3Dv2}~\cite{reizenstein2021common}} &
    \multicolumn{4}{c}{\textbf{ScanNet++}~\cite{yeshwanth2023scannet++}} &
    \multicolumn{4}{c}{\textbf{DL3DV}~\cite{ling2024dl3dv}} 
    \\
    \cmidrule(lr){3-6} \cmidrule(lr){7-10} \cmidrule(lr){11-14} \cmidrule(lr){15-18}
    & & PSNR$_\uparrow$ & @5°$_\uparrow$ & @15°$_\uparrow$ & @30°$_\uparrow$
      & PSNR$_\uparrow$ & @5°$_\uparrow$ & @15°$_\uparrow$ & @30°$_\uparrow$
      & PSNR$_\uparrow$ & @5°$_\uparrow$ & @15°$_\uparrow$ & @30°$_\uparrow$
      & PSNR$_\uparrow$ & @5°$_\uparrow$ & @15°$_\uparrow$ & @30°$_\uparrow$ \\
    \midrule

    \multirow{2}{*}{RE10K~\cite{zhou2018stereo}} 
    & \cellcolor{tablered2}VGGT* & \cellcolor{gray!10}/ & \cellcolor{yellow}0.4 & \cellcolor{yellow}8.4 & \cellcolor{yellow}22.5 & \cellcolor{gray!10}/ & \cellcolor{yellow}0.1 & \cellcolor{yellow}3.7 & \cellcolor{yellow}15.5 & \cellcolor{gray!10}/ & \cellcolor{yellow}0.6 & \cellcolor{yellow}10.0 & \cellcolor{yellow}30.7 & \cellcolor{gray!10}/ & \cellcolor{yellow}17.8 & \cellcolor{yellow}50.9 & \cellcolor{yellow}69.4 \\
    & \cellcolor{tablegreen}E-RayZer & \cellcolor{yellow}17.2 & \cellcolor{yellow}\underline{1.8} & \cellcolor{yellow}\underline{16.9} & \cellcolor{yellow}\underline{34.0} & \cellcolor{yellow}19.1 & \cellcolor{yellow}\underline{0.6} & \cellcolor{yellow}\underline{8.3} & \cellcolor{yellow}\underline{26.0} & \cellcolor{yellow}17.5 & \cellcolor{yellow}\underline{1.1} & \cellcolor{yellow}\underline{13.3} & \cellcolor{yellow}\underline{37.3} & \cellcolor{yellow}17.3 & \cellcolor{yellow}\underline{21.2} & \cellcolor{yellow}\underline{55.0} & \cellcolor{yellow}\underline{72.7} \\
    \midrule

    \multirow{2}{*}{DL3DV~\cite{ling2024dl3dv}} 
    & \cellcolor{tablered2}VGGT* & \cellcolor{gray!10}/ & \cellcolor{orange}14.3 & \cellcolor{orange}54.5 & \cellcolor{orange}75.7 & \cellcolor{gray!10}/ & \cellcolor{orange}16.0 & \cellcolor{orange}64.3 & \cellcolor{orange}82.1 & \cellcolor{gray!10}/ & \cellcolor{orange}6.7 & \cellcolor{orange}39.8 & \cellcolor{orange}71.5 & \cellcolor{gray!10}/ & \cellcolor{tablered}79.6 & \cellcolor{tablered}94.2 & \cellcolor{tablered}97.1 \\
    & \cellcolor{tablegreen}E-RayZer & \cellcolor{orange}20.5 & \cellcolor{orange}\underline{20.7} & \cellcolor{tablered}\underline{57.8} & \cellcolor{tablered}69.6 &
    \cellcolor{orange}22.9 & \cellcolor{orange}\underline{19.1} & \cellcolor{orange}61.8 & \cellcolor{orange}78.8 & \cellcolor{orange}20.1 & \cellcolor{tablered}\underline{7.7} & \cellcolor{orange}33.6 & \cellcolor{orange}63.0 &
    \cellcolor{tablered}20.3 & \cellcolor{tablered}72.0 & \cellcolor{tablered}88.4 & \cellcolor{tablered}93.5 \\
    \midrule


    \multirow{2}{*}{7-dataset Mix}
    & \cellcolor{tablered2}VGGT* & \cellcolor{gray!10}/ & \cellcolor{tablered}28.8 & \cellcolor{tablered}67.3 & \cellcolor{tablered}84.4 & \cellcolor{gray!10}/ & \cellcolor{tablered}43.4 & \cellcolor{tablered}83.5 & \cellcolor{tablered}91.8 & \cellcolor{gray!10}/ & \cellcolor{tablered}13.1 & \cellcolor{tablered}54.8 & \cellcolor{tablered}78.5 & \cellcolor{gray!10}/ & \cellcolor{orange}66.1 & \cellcolor{orange}88.9 & \cellcolor{orange}95.6 \\
    & \cellcolor{tablegreen}E-RayZer & \cellcolor{tablered} 20.6 & \cellcolor{tablered}24.6 & \cellcolor{orange} 56.1 & \cellcolor{orange} 69.2 &
      \cellcolor{tablered} 24.3 & \cellcolor{tablered} 30.3 & \cellcolor{tablered} 74.2 & \cellcolor{tablered} 83.7 &
      \cellcolor{tablered} 20.7 & \cellcolor{orange} 5.7 & \cellcolor{tablered} 34.8 & \cellcolor{tablered} 63.7 &
      \cellcolor{orange} 19.7 & \cellcolor{orange} 59.9 & \cellcolor{orange} 82.9 & \cellcolor{orange} 90.2 \\
    \bottomrule
  \end{tabular}}
  \vspace{-0.2in}
\end{table*}

\subsubsection{Probing Representations on Downstream Tasks}
\label{sec:exp_downstream}

\noindent \textbf{Baselines and Setups.} To further assess the spatial awareness, we probe and compare the feature representations of \ours{} against widely-used vision encoders: DINO series~\cite{oquab2023dinov2, simeoni2025dinov3}, CroCo v2~\cite{weinzaepfel2023croco}, VideoMAE V2~\cite{wang2023videomae}, Perception Encoder~\cite{bolya2025perception}, and RayZer~\cite{jiang2025rayzer}. We only use the backbones and train the prediction heads \textbf{from scratch}. We compare performance under both frozen-backbone and full-finetuning settings on downstream tasks, including:

\hspace*{0.05in}\textbullet~\emph{Multi-view Depth and Pose Estimation (3D Tasks).} For depth estimation, we apply a DPT head~\cite{ranftl2021vision} on top of the backbones. 
Single-view backbones lack multi-view correspondence and thus reduce to monocular depth estimation.
For pose estimation, we attach VGGT’s~\cite{wang2025vggt} camera head to each backbone, using either the class token or averaged patch features as camera tokens. 
These tokens are aggregated across views via transformer layers, enabling even single-view models to reason over multi-view geometry. We note that the camera heads of RayZer and \ours{} from their pre-training stage are \textit{not} used for fairness.

\hspace*{0.05in}\textbullet~\emph{Pairwise Flow Estimation (2.5D Task).} We consider backbones that encode binocular geometry, including CroCo v2~\cite{weinzaepfel2023croco}, VideoMAE V2~\cite{wang2023videomae}, RayZer~\cite{jiang2025rayzer}, and \ours{}. We follow the settings of UFM~\cite{zhang2025ufm}.

\noindent \textbf{Results on 3D Downstream Tasks.} Tab.~\ref{tab:comparison_pose_depth} shows that \ours{} achieves the best performance across all datasets and settings, demonstrating strong 3D-awareness in its feature representations. Under the frozen-backbone setting, \ours{} notably outperforms all baselines. With full finetuning, \ours{} further improves across all metrics, surpassing RayZer~\cite{jiang2025rayzer} and VideoMAE V2~\cite{wang2023videomae} by a large margin. The consistently strong results highlight the generalization ability of its geometrically grounded representations, showing its potential as a spatial visual pre-training framework. 

\noindent \textbf{Results on Pairwise Flow Estimation.} Tab.~\ref{tab:comparison_flow} shows that \ours{} achieves competitive performance on pairwise flow prediction, closely following RayZer~\cite{jiang2025rayzer}, despite not being trained directly for tasks that optimize image correspondences (\eg, masked image modeling in CroCo v2~\cite{weinzaepfel2023croco} and VideoMAE V2~\cite{wang2023videomae}, or view interpolation in RayZer). Compared to \ours{}, RayZer holds a slight advantage due to its implicit 3D formulation, naturally suited for low-level motion estimation. Nevertheless, \ours{} outperforms other baselines, demonstrating that its explicit 3D representation learning captures meaningful spatial correspondences even for the 2.5D task.




\vspace{-0.01in}
\subsection{Ablation Study}
\label{sec:exp_ablation}
\vspace{-0.01in}

\noindent \textbf{Data Mixing / Scaling.} We investigate the behavior of self-supervised \ours{} and supervised VGGT* (Sec.~\ref{sec: exp_ours_and_vggt}) under varying data scales and quality. In Tab.~\ref{tab:data_mixing_ablation}, \ours{} and VGGT* demonstrate a similar scaling behavior: training on data with broader distributions improves generalization (\eg, models trained on 7 datasets outperform those trained on DL3DV alone). However, reducing the sampling frequency of a particular domain slightly degrades performance on its corresponding test set (\eg, 7-dataset models perform worse on DL3DV than DL3DV-only models), a trend consistently observed in prior work~\cite{xie2023doremi, ye2024data, foroutan2025revisiting}. Besides, data quality also plays a key role, as training on DL3DV yields better results than that on RE10K. 

Moreover, again, the self-supervised model (\ours{}) achieves performance on par with the supervised VGGT* (while VGGT* holds advantage when trained on large data), demonstrating that large-scale self-supervision alone can yield geometrically grounded 3D understanding.
This result underscores that data diversity and quality, rather than explicit 3D supervision, are the true drivers of scalability in large 3D vision models. Together, these results highlight the great potential of self-supervised 3D learning when scaled to internet-scale data, and provide valuable guidance for future data selection and curation strategies.

\begin{table}[t]
\centering
\small
\caption{\textbf{Ablation on Curriculum Learning.} We compare four curriculum strategies when training \ours{} on DL3DV (top) and a seven-dataset mixture (bottom). The proposed visual-overlap-based curriculum consistently outperforms baselines.}
\vspace{-0.1in}
\label{tab:ablation_curriculum}
\footnotesize
\renewcommand{\arraystretch}{0.98}
\resizebox{\linewidth}{!}{
\begin{tabular}{clc|ccc}
    \toprule
    & \textbf{Curriculum Variant} & \textbf{PSNR}$_\uparrow$ & \textbf{RPA@5°}$_\uparrow$ & \textbf{RPA@15°}$_\uparrow$ & \textbf{RPA@30°}$_\uparrow$ \\
    \midrule
    \parbox[t]{2mm}{\multirow{4}{*}{\rotatebox[origin=c]{90}{\footnotesize DL3DV}}}
        & No Curriculum          & 16.1 & 4.0  & 27.8 & 47.2  \\
        & Frame Interval         & 19.8 & 56.1 & 79.3 & 86.0  \\
        & Semantic Overlap       & \textbf{20.4} & \textbf{73.2} & \textbf{88.7} & \textbf{93.7}  \\
        & Geometric Overlap      & \underline{20.3} & \underline{72.0} & \underline{88.4} & \underline{93.5} \\
    \midrule
    \parbox[t]{2mm}{\multirow{4}{*}{\rotatebox[origin=c]{90}{\footnotesize 7-dataset}}}
        & No Curriculum          & 15.9 & 2.1  & 21.6 & 40.7 \\
        & Frame Interval         & 19.1 & 43.8 & 72.1 & 82.9 \\
        & Semantic Overlap       & \textbf{19.7} & \underline{58.7} & \underline{81.0} & \underline{89.8} \\
        & Geometric Overlap      & \textbf{19.7} & \textbf{59.9} & \textbf{82.9} & \textbf{90.2} \\
    \bottomrule
\end{tabular}
}
\vspace{-0.23in}
\end{table}
 
\noindent \textbf{Curriculum Learning.}  In Tab.~\ref{tab:ablation_curriculum}, we compare against two baselines with (1) no curriculum, and (2) a frame-interval-based curriculum, where frame intervals are specified for each dataset. Across two training regimes (\ie, DL3DV-only and the seven-dataset mixture), the proposed visual-overlap curricula consistently outperform both baselines, with the two variants performing comparably. These results demonstrate that our fine-grained curriculum strategy significantly improves self-supervised pose estimation and reconstruction, while eliminating the need for manual tuning for each training dataset and benefiting scaling.
\section{Discussion}
We propose \ours{}, a multi-view 3D model that learns geometrically grounded representations via self-supervised 3D reconstruction. While our results are promising, we identify several limitations and directions for future work.

\ours{} currently operates on static scenes, limiting its training to existing static datasets -- high-quality static-scene data remains scarce beyond these benchmarks. Extending it to dynamic scenes would enable learning from generic videos and is a promising direction. Additionally, our learning curriculum generally assumes continuous video frames with fairly uniform camera motion; sparse images or frames with drastic viewpoint changes may reduce its effectiveness.

Despite these limitations, \ours{} outperforms prior self-supervised methods and is competitive with supervised approaches. Extensive experiments show that \ours{} pre-training benefits supervised models and other 3D downstream tasks, establishing it as a scalable 3D-aware visual pre-training framework.

\newpage

\noindent \textbf{Acknowledgements.} The work is partially done during Qitao Zhao's internship at Adobe Research. We thank Zhengqi Li for insightful advice. We also thank Frédéric Fortier-Chouinard, Jiashun Wang, Yanbo Xu, Zihan Wang, and members of the Physical Perception Lab for helpful discussions.

This work was also supported by Intelligence Advanced Research Projects Activity (IARPA) via Department of Interior/Interior Business Center (DOI/IBC) contract number 140D0423C0074. The U.S. Government is authorized to reproduce and distribute reprints for Governmental purposes notwithstanding any copyright annotation thereon. Disclaimer: The views and conclusions contained herein are those of the authors and should not be interpreted as necessarily representing the official policies or endorsements, either expressed or implied, of IARPA, DOI/IBC, or the U.S. Government.

\clearpage
\setcounter{page}{1}
\maketitlesupplementary

\appendix

\section*{Overview}
This supplementary material is organized as follows:
\begin{itemize}
\item Section~\ref{sec:supp_imple_details}: Additional implementation details.
\item Section~\ref{sec:supp_finetune_details}: Details on supervised finetuning.
\item Section~\ref{sec:supp_curriculum_details}: Additional details on curriculum learning ablations.
\item Section~\ref{sec:supp_pose_supervised_baseline}: Analysis of \ours{} trained with pose supervision.
\item Section~\ref{sec:supp_pretraining}: Additional results where \ours{} is used as pre-training for the VGGT* model, with comparisons to RayZer~\cite{jiang2025rayzer}.
\item Section~\ref{sec:supp_training_data}: Further analysis of the training data.
\item Section~\ref{sec:supp_qualitative_comparison}: Extended qualitative comparisons with baseline methods.
\end{itemize}





\section{Additional Implementation Details}
\label{sec:supp_imple_details}

This section includes more implementation details.

\noindent \textbf{Training.} \ours{} is trained on 8 A100 GPUs with a global batch size of 192 (24 per GPU) for 152K iterations, taking approximately 198 hours. During the first 86K iterations, the learning curriculum progresses linearly along different sequence-sampling metrics -- geometric (default) and semantic visual overlap, as well as frame interval -- as described in Sec.~\ref{sec:exp_ablation}. The learning rate schedule includes a 3K-iteration linear warm-up to a peak of 4e\text{-}4, followed by cosine decay to zero til the end of training. We use the AdamW optimizer ($\beta_{1}$=0.9, $\beta_{2}$=0.95) with gradient clipping at 1.0, and skip optimization steps when the gradient norm exceeds 5.0 before clipping.

For our 7-dataset model (Sec.~\ref{sec:exp_setup}), we train on a mixture of datasets with the following sampling ratios:
DL3DV~\cite{ling2024dl3dv}: 1.0, CO3Dv2~\cite{reizenstein2021common}: 0.25, RealEstate10K~\cite{zhou2018stereo}: 0.5, MVImgNet~\cite{yu2023mvimgnet}: 0.25, ARKitScenes~\cite{baruch2021arkitscenes}: 0.5, WildRGB-D~\cite{xia2024rgbd}: 0.25, and ACID~\cite{liu2021infinite}: 0.5. These ratios follow a simple heuristic: we downweight object-centric datasets and assign a slightly larger weight to DL3DV, which offers the most diverse and high-quality samples.

Experiments on supervised finetuning are conducted on 8 A100 GPUs as well, but with a smaller global batch size of 96. The finetuning stage runs for 50K iterations.

\noindent \textbf{Architecture.} \ours{} uses a patch size of 16 and an image resolution of 256. As described in Sec.~\ref{sec:architecture}, we replace RayZer’s~\cite{jiang2025rayzer} vanilla global attention with VGGT’s~\cite{wang2025vggt} local-global alternating transformer layers for both pose estimation ($f_{\boldsymbol{\theta}}^{\text{cam}}$) and scene reconstruction ($f_{\boldsymbol{\psi}'}^{\text{scene}}$).
Both modules use 8 layers, each composed of one global attention layer and one frame-attention layer. Our feature dimension is 768, and we use 12 attention heads. For image and Plücker ray map tokenization, as well as for the Gaussian decoder ($f_{\boldsymbol{\omega}}^{\text{gauss}}$), we simply use a single linear layer.
 
For a fair comparison with RayZer, all RayZer models used in this paper are trained with our proposed curriculum and the improved architecture.

\noindent \textbf{Evaluation.} For pose estimation and novel-view synthesis, we use fixed sequence lengths for the test sequences of each dataset and sample views with equal temporal spacing. Following RayZer, we ensure that the first and last images of each sequence are always included in the reference set. The sequence lengths are as follows: WildRGB-D~\cite{xia2024rgbd}: 96 (Tab.~\ref{tab:comparison_pose_and_nvs}) and 192 (Tab.~\ref{tab:comparison_with_vggt}), ScanNet++~\cite{yeshwanth2023scannet++}: 48, DL3DV~\cite{ling2024dl3dv}: 96, RealEstate10K~\cite{zhou2018stereo}: 256, CO3Dv2~\cite{reizenstein2021common}: 96, 7-Scenes~\cite{shotton2013scene}: 256, Cambridge Landmarks~\cite{kendall2015posenet}: 96, BlendedMVS~\cite{yao2020blendedmvs}: 24, and NAVI~\cite{jampani2024navi}: 24.
For (training and) evaluating pairwise flow prediction on StaticThings3D~\cite{schroppel2022benchmark}, we adopt the pre-computed image pairs provided by the DUSt3R~\cite{wang2023DUSt3R} GitHub repository.

\begin{table*}[t]
  \centering
  \caption{\textbf{Comparison with a Pose-supervised Baseline on Novel-view Synthesis (NVS) and Pose Estimation.} We report PSNR for NVS and RPA$_\uparrow$@5°/15°/30° for pose estimation. While the pose-supervised baseline generally outperforms the self-supervised model on coarse pose accuracy (RPA$_\uparrow$@15°/30°), its novel-view synthesis quality is consistently lower.}
  \vspace{-0.12in}
  \label{tab:supp_comparison_pose_supervised}
  \resizebox{\textwidth}{!}{
  \begin{tabular}{lccccccccccccc}
    \toprule
    \multirow{2}{*}{\textbf{Method}} & \multirow{2}{*}{\textbf{Training Data}}  &
    \multicolumn{4}{c}{\textbf{NAVI}~\cite{jampani2024navi}} &
    \multicolumn{4}{c}{\textbf{ScanNet++}~\cite{yeshwanth2023scannet++}} &
    \multicolumn{4}{c}{\textbf{DL3DV}~\cite{ling2024dl3dv}} \\
    \cmidrule(lr){3-6} \cmidrule(lr){7-10} \cmidrule(lr){11-14}
    & & PSNR$_\uparrow$ & @5°$_\uparrow$ & @15°$_\uparrow$ & @30°$_\uparrow$
    & PSNR$_\uparrow$ & @5°$_\uparrow$ & @15°$_\uparrow$ & @30°$_\uparrow$
    & PSNR$_\uparrow$ & @5°$_\uparrow$ & @15°$_\uparrow$ & @30°$_\uparrow$ \\
    \midrule
    Pose-sup. Baseline & \multirow{2}{*}{DL3DV~\cite{ling2024dl3dv}} & 
    13.4 & 12.8 & 51.1 & \textbf{72.5} &
    16.7 & 4.4  & \textbf{33.7} & \textbf{64.5} &
    15.0 & \textbf{78.1} & \textbf{94.7} & \textbf{97.8} \\
    \ours{} (ours) & &
    \textbf{20.5} & \textbf{20.7} & \textbf{57.8} & 69.6 & 
    \textbf{20.1} & \textbf{7.7}  & 
    33.6 & 63.0 & 
    \textbf{20.3} & 72.0 & 88.4 & 93.5 \\
    \midrule
    Pose-sup. Baseline & \multirow{2}{*}{7 datasets} &
    13.5 & 18.9 & \textbf{61.6} & \textbf{80.6} &
    17.3 & \textbf{6.4}  & \textbf{35.7} & \textbf{67.4} &
    14.9 & 53.0 & \textbf{85.0} & \textbf{93.2} \\
    \ours\ (ours) & &
    \textbf{20.6} & \textbf{24.6} & 56.1 & 69.2 & 
    \textbf{20.7} & 5.7  & 34.8 & 63.7 & 
    \textbf{19.7} & \textbf{59.9} & 82.9 & 90.2 \\
    \bottomrule
  \end{tabular}}
\end{table*}

\begin{table*}[t]
  \centering
  \caption{\textbf{Comparison with RayZer~\cite{jiang2025rayzer} as a Pre-trained Backbone.} The top block reports results for models trained on DL3DV~\cite{ling2024dl3dv}, and the bottom block reports results for models trained on a mixture of seven datasets. Note that pre-training and supervised finetuning are performed on the same data (\ie, DL3DV or the 7-dataset mixture). We report pose accuracy RPA$_\uparrow@5^\circ$/$15^\circ$. Models are labeled as \textcolor[rgb]{0.35,0.8,0.35}{self-supervised} or \textcolor{tablered2!250}{supervised}. VGGT* denotes our re-implementation with \ours{}'s pairwise camera head. The top-three results are color-ranked from red to yellow. \ours{} provides stronger pre-training than RayZer.}
  \vspace{-0.12in}
  \label{tab:supp_comparison_with_vggt}
  \resizebox{\textwidth}{!}{
  \renewcommand{\arraystretch}{1.02}
  \begin{tabular}{lp{3.4cm}@{}cc|cccccccccccccccc}
    \toprule
     & \multirow{2}{*}{\textbf{Method}} &
    \multicolumn{2}{c|}{\textbf{DL3DV~\cite{ling2024dl3dv}}} &
    \multicolumn{2}{c}{\textbf{RE10K~\cite{zhou2018stereo}}} &
    \multicolumn{2}{c}{\textbf{CO3Dv2~\cite{reizenstein2021common}}} &
    \multicolumn{2}{c}{\textbf{WildRGB-D~\cite{xia2024rgbd}}} &
    \multicolumn{2}{c}{\textbf{7-Scenes~\cite{shotton2013scene}}} &
    \multicolumn{2}{c}{\textbf{CamLand~\cite{kendall2017geometric}}} &
    \multicolumn{2}{c}{\textbf{BlendedMVS~\cite{yao2020blendedmvs}}} &
    \multicolumn{2}{c}{\textbf{NAVI~\cite{jampani2024navi}}} &
    \multicolumn{2}{c}{\textbf{ScanNet++~\cite{yeshwanth2023scannet++}}} \\
    \cmidrule(lr){3-4} \cmidrule(lr){5-6} \cmidrule(lr){7-8} \cmidrule(lr){9-10}
    \cmidrule(lr){11-12} \cmidrule(lr){13-14} \cmidrule(lr){15-16} \cmidrule(lr){17-18} \cmidrule(lr){19-20}
    & & @5°$_\uparrow$ & @15°$_\uparrow$ 
      & @5°$_\uparrow$ & @15°$_\uparrow$
      & @5°$_\uparrow$ & @15°$_\uparrow$
      & @5°$_\uparrow$ & @15°$_\uparrow$
      & @5°$_\uparrow$ & @15°$_\uparrow$
      & @5°$_\uparrow$ & @15°$_\uparrow$
      & @5°$_\uparrow$ & @15°$_\uparrow$
      & @5°$_\uparrow$ & @15°$_\uparrow$
      & @5°$_\uparrow$ & @15°$_\uparrow$ \\
    \midrule

    \multirow{5}{*}{\rotatebox[origin=c]{90}{\parbox{2.5cm}{\centering DL3DV}}}
    & \solidcell{tablegreen}{RayZer~\cite{jiang2025rayzer}}
    & \cellcolor{white} 0.0 & 0.6 
    & 0.0 & 0.2 
    & 0.0 & 0.6 
    & 0.0 & 0.0
    & 0.0 & 0.2 
    & 0.0 & 0.3 
    & 0.0 & 0.5 
    & 0.0 & 0.6 
    & 0.0 & 0.7 \\

    & \solidcell{tablegreen}{\ours{} (ours)}
    & \cellcolor{white} 72.0 & 88.4 
    & \cellcolor{yellow} 83.0 & 96.8 
    & \cellcolor{yellow} 19.1 & 61.8 
    & \cellcolor{orange} 51.1 & \cellcolor{yellow} 82.3 
    & \cellcolor{orange} 38.8 & 78.0 
    & \cellcolor{yellow} 18.1 & \cellcolor{yellow} 62.9 
    & \cellcolor{yellow} 22.9 & \cellcolor{orange} 46.8 
    & \cellcolor{yellow} 20.7 & \cellcolor{yellow} 57.8 
    & \cellcolor{yellow} 7.7  & 33.6 \\

    & \solidcell{tablered2}{VGGT*}
    & \cellcolor{yellow} 79.6 & \cellcolor{yellow} 94.2
    & \cellcolor{white} 80.4 & \cellcolor{yellow} 97.9
    & 16.0 & \cellcolor{yellow} 64.3
    & 32.5 & 76.2
    & 34.7 & \cellcolor{tablered} 83.6
    & 11.1 & 49.8
    & 17.0 & 42.8
    & 14.3 & 54.5
    & 6.7  & \cellcolor{yellow} 39.8 \\

    & \tikz[baseline=(n.base)]{
        \node[
          inner sep=0.4ex, outer sep=0pt,
          minimum width=\linewidth,
          minimum height=1.6em,
          shade,
          left color=tablegreen!110,
          right color=tablered2!110
        ] (n) {\makebox[\linewidth][l]{\strut RayZer→VGGT*}};
      }
    & \cellcolor{orange} 84.4 & \cellcolor{orange} 95.3
    & \cellcolor{tablered} 85.7 & \cellcolor{tablered} 98.4
    & \cellcolor{orange} 24.9 & \cellcolor{orange} 71.2
    & \cellcolor{yellow} 43.9 & \cellcolor{orange} 86.4
    & \cellcolor{yellow} 38.0 & \cellcolor{tablered} 83.6
    & \cellcolor{orange} 27.3 & \cellcolor{orange} 73.0
    & \cellcolor{orange} 24.0 & \cellcolor{yellow} 45.8
    & \cellcolor{orange} 25.5 & \cellcolor{orange} 58.3
    & \cellcolor{orange} 12.2 & \cellcolor{orange} 49.6 \\

    & \tikz[baseline=(n.base)]{
        \node[
          inner sep=0.4ex, outer sep=0pt,
          minimum width=\linewidth,
          minimum height=1.6em,
          shade,
          left color=tablegreen!110,
          right color=tablered2!110
        ] (n) {\makebox[\linewidth][l]{\strut \ours{}→VGGT*}};
      }
    & \cellcolor{tablered} 87.3 & \cellcolor{tablered} 96.6
    & \cellcolor{orange} 85.3 & \cellcolor{tablered} 98.4
    & \cellcolor{tablered} 25.3 & \cellcolor{tablered} 72.2
    & \cellcolor{tablered} 56.2 & \cellcolor{tablered} 91.4
    & \cellcolor{tablered} 43.8 & \cellcolor{yellow} 82.8
    & \cellcolor{tablered} 30.2 & \cellcolor{tablered} 75.6
    & \cellcolor{tablered} 29.2 & \cellcolor{tablered} 52.2
    & \cellcolor{tablered} 26.9 & \cellcolor{tablered} 64.3
    & \cellcolor{tablered} 14.3 & \cellcolor{tablered} 53.8 \\
    \midrule

    \multirow{5}{*}{\rotatebox[origin=c]{90}{\parbox{2.5cm}{\centering 7 datasets}}}
    & \solidcell{tablegreen}{RayZer~\cite{jiang2025rayzer}}
    & \cellcolor{white} 0.0 & 1.9 
    & 0.0 & 0.9 
    & 0.0 & 1.6
    & 0.0 & 1.1 
    & 0.0 & 2.0 
    & 0.0 & 0.6 
    & 0.0 & 1.6 
    & 0.0 & 1.6 
    & 0.0 & 0.9 \\

    & \solidcell{tablegreen}{\ours{} (ours)}
    & \cellcolor{white} 59.9 & 82.9 
    & 84.1 & 97.5 
    & 30.3 & 74.2 
    & 63.1 & 85.3 
    & 26.0 & 76.5 
    & 9.8  & 47.3 
    & \cellcolor{yellow} 22.3 & 45.5 
    & 24.6 & 56.1 
    & 5.7  & 34.8 \\

    & \solidcell{tablered2}{VGGT*}
    & \cellcolor{yellow} 66.1 & \cellcolor{yellow} 88.9
    & \cellcolor{yellow} 85.2 & \cellcolor{yellow} 98.5
    & \cellcolor{yellow} 43.4 & \cellcolor{yellow} 83.5
    & \cellcolor{yellow} 76.8 & \cellcolor{yellow} 96.0
    & \cellcolor{yellow} 31.1 & \cellcolor{yellow} 78.0
    & \cellcolor{yellow} 22.9 & \cellcolor{tablered} 66.3
    & 19.0 & \cellcolor{yellow} 49.9
    & \cellcolor{yellow} 28.8 & \cellcolor{yellow} 67.3
    & \cellcolor{yellow} 13.1 & \cellcolor{yellow} 54.8 \\

    & \tikz[baseline=(n.base)]{
        \node[
          inner sep=0.4ex, outer sep=0pt,
          minimum width=\linewidth,
          minimum height=1.6em,
          shade,
          left color=tablegreen!110,
          right color=tablered2!110
        ] (n) {\makebox[\linewidth][l]{\strut RayZer→VGGT*}};
      }
    & \cellcolor{orange} 72.8 & \cellcolor{orange} 91.7
    & \cellcolor{orange} 88.1 & \cellcolor{orange} 98.6
    & \cellcolor{orange} 53.8 & \cellcolor{orange} 85.1
    & \cellcolor{orange} 81.5 & \cellcolor{orange} 96.3
    & \cellcolor{orange} 37.7 & \cellcolor{orange} 84.9
    & \cellcolor{orange} 28.3 & \cellcolor{orange} 65.7
    & \cellcolor{orange} 24.3 & \cellcolor{orange} 52.7
    & \cellcolor{orange} 34.6 & \cellcolor{orange} 70.4
    & \cellcolor{orange} 15.0 & \cellcolor{orange} 58.7 \\

    & \tikz[baseline=(n.base)]{
        \node[
          inner sep=0.4ex, outer sep=0pt,
          minimum width=\linewidth,
          minimum height=1.6em,
          shade,
          left color=tablegreen!110,
          right color=tablered2!110
        ] (n) {\makebox[\linewidth][l]{\strut \ours{}→VGGT*}};
      }
    & \cellcolor{tablered} 78.8 & \cellcolor{tablered} 92.8
    & \cellcolor{tablered} 91.0 & \cellcolor{tablered} 99.1
    & \cellcolor{tablered} 58.9 & \cellcolor{tablered} 86.3
    & \cellcolor{tablered} 86.4 & \cellcolor{tablered} 96.7
    & \cellcolor{tablered} 42.7 & \cellcolor{tablered} 88.3
    & \cellcolor{tablered} 35.2 & \cellcolor{yellow} 64.4
    & \cellcolor{tablered} 31.5 & \cellcolor{tablered} 57.7
    & \cellcolor{tablered} 41.5 & \cellcolor{tablered} 73.7
    & \cellcolor{tablered} 22.0 & \cellcolor{tablered} 65.2 \\

    \bottomrule
  \end{tabular}}
  \vspace{-0.14in}
\end{table*}

\begin{table*}[t]
  \centering
  \small
  \caption{\textbf{Additional Results on Data Mixing and Scaling.} We train \ours{} with different combinations of datasets. Compared to Tab.~\ref{tab:data_mixing_ablation}, we additionally include SpatialVID~\cite{wang2025spatialvid}, a large in-the-wild video dataset. Results are color-ranked from red to yellow. Mixing datasets improves distribution coverage, whereas simply using larger datasets does not necessarily yield better performance -- both diversity and data quality play critical roles.}
  \vspace{-0.12in}
  \label{tab:supp_data_mixing_ablation}
  \renewcommand{\arraystretch}{0.98}
  \resizebox{\textwidth}{!}{
  \begin{tabular}{lccccccccccccccccc}
    \toprule
    \multirow{2}{*}{\textbf{Training Data}} &
    \multirow{2}{*}{\textbf{\# Seq.}} &
    \multicolumn{4}{c}{\textbf{NAVI}~\cite{jampani2024navi}} &
    \multicolumn{4}{c}{\textbf{CO3Dv2}~\cite{reizenstein2021common}} &
    \multicolumn{4}{c}{\textbf{ScanNet++}~\cite{yeshwanth2023scannet++}} &
    \multicolumn{4}{c}{\textbf{DL3DV}~\cite{ling2024dl3dv}} 
    \\
    \cmidrule(lr){3-6} \cmidrule(lr){7-10} \cmidrule(lr){11-14} \cmidrule(lr){15-18}
    & & PSNR$_\uparrow$ & @5°$_\uparrow$ & @15°$_\uparrow$ & @30°$_\uparrow$
      & PSNR$_\uparrow$ & @5°$_\uparrow$ & @15°$_\uparrow$ & @30°$_\uparrow$
      & PSNR$_\uparrow$ & @5°$_\uparrow$ & @15°$_\uparrow$ & @30°$_\uparrow$
      & PSNR$_\uparrow$ & @5°$_\uparrow$ & @15°$_\uparrow$ & @30°$_\uparrow$ \\
    \midrule

    RE10K~\cite{zhou2018stereo} & 66K & 17.2 & \cellcolor{yellow} 1.8 & \cellcolor{yellow} 16.9 & \cellcolor{yellow} 34.0 & 19.1 & \cellcolor{yellow} 0.6 & \cellcolor{yellow} 8.3 & \cellcolor{yellow} 26.0 & 17.5 & \cellcolor{yellow} 1.1 & \cellcolor{yellow} 13.3 & \cellcolor{yellow} 37.3 & \cellcolor{yellow}17.3 & \cellcolor{yellow} 21.2 & \cellcolor{yellow} 55.0 & \cellcolor{yellow} 72.7 \\

    SpatialVID~\cite{wang2025spatialvid} & 100K & \cellcolor{yellow} 17.9 & 0.7 & 11.2 & 26.4 &
    \cellcolor{yellow} 19.9 & 0.2 & 5.7 & 20.9 & \cellcolor{yellow} 18.0 & 0.3 & 6.7 & 26.0 &
    17.2 & 11.4 & 36.6 & 56.0 \\

    DL3DV~\cite{ling2024dl3dv} & 10K & \cellcolor{orange} 20.5 & \cellcolor{orange} 20.7 & \cellcolor{tablered} 57.8 & \cellcolor{tablered} 69.6 &
    \cellcolor{orange} 22.9 & \cellcolor{orange} 19.1 & \cellcolor{orange} 61.8 & \cellcolor{orange} 78.8 & \cellcolor{orange} 20.1 & \cellcolor{tablered} 7.7 & \cellcolor{orange} 33.6 & \cellcolor{orange} 63.0 &
    \cellcolor{tablered} 20.3 & \cellcolor{tablered} 72.0 & \cellcolor{tablered} 88.4 & \cellcolor{tablered} 93.5 \\

    7-dataset Mix & 352K & \cellcolor{tablered} 20.6 & \cellcolor{tablered}24.6 & \cellcolor{orange} 56.1 & \cellcolor{orange} 69.2 &
      \cellcolor{tablered} 24.3 & \cellcolor{tablered} 30.3 & \cellcolor{tablered} 74.2 & \cellcolor{tablered} 83.7 &
      \cellcolor{tablered} 20.7 & \cellcolor{orange} 5.7 & \cellcolor{tablered} 34.8 & \cellcolor{tablered} 63.7 &
      \cellcolor{orange} 19.7 & \cellcolor{orange} 59.9 & \cellcolor{orange} 82.9 & \cellcolor{orange} 90.2 \\
    \bottomrule
  \end{tabular}}
\end{table*}

\begin{figure*}[t]
  \centering
  \includegraphics[width=\textwidth]{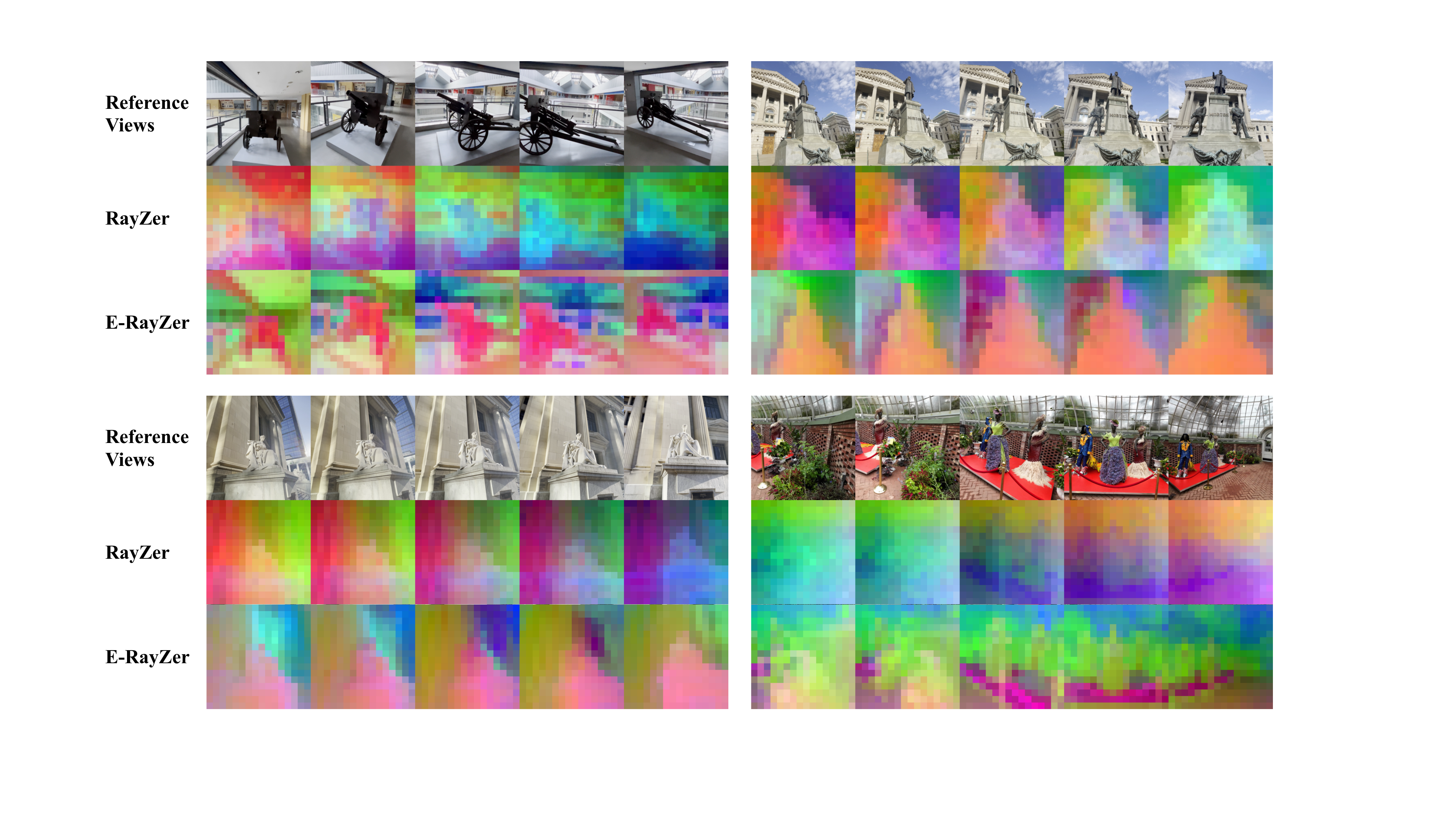}
  \vspace{-0.27in}
  \caption{\textbf{Additional Visual Comparison with RayZer~\cite{jiang2025rayzer} on Learned Features.} We visualize feature maps using their top-three PCA components. The features produced by \ours{} exhibit stronger and more spatially consistent patterns that align well with the underlying scene structure, whereas RayZer’s features show noticeable color shifts across frames.}
  \label{fig:supp_vis_feature}
  \vspace{-0.05in}
\end{figure*}

\section{More Details on Supervised Finetuning}
\label{sec:supp_finetune_details}

Here we provide additional details on the supervised finetuning experiments in Sec.~\ref{sec:exp_pretraining}.

\noindent \textbf{Supervised Finetuning with \ours{}.} E-RayZer’s backbone does not distinguish between the first view and the other views in the input, as it adopts a pairwise pose estimation strategy (see Sec.~\ref{sec:architecture}). In contrast, supervised pose estimation typically assumes a first-view coordinate frame (\eg, DUSt3R~\cite{wang2023DUSt3R} and VGGT~\cite{wang2025vggt}). To incorporate this inductive bias into our backbone, we introduce an additional camera token dedicated to the first image (in addition to the existing learned camera token) and train it from scratch. The camera tokens are processed by \ours{}’s pose estimation module ($f_{\boldsymbol{\theta}}^{\text{cam}}$) and subsequently passed to VGGT’s camera head for supervised pose estimation. For depth estimation and pairwise flow prediction, the DPT head takes as input the intermediate feature maps generated by the Gaussian-based scene reconstruction module ($f_{\boldsymbol{\psi}'}^{\text{scene}}$). For \ours{} and all other baselines, the DPT head uses four feature maps extracted from equally spaced transformer layers. Note that our Gaussian-based scene reconstruction module takes the predicted reference-view Plücker ray maps as input, but only in the pose and depth estimation experiments are the predicted camera poses supervised. For pairwise flow prediction, the predicted poses produced by the pose head remain unsupervised to ensure a fair comparison with other baselines.

\noindent \textbf{Details on Other Baselines.} For baselines that use different spatial or temporal patch sizes (\eg, \ours{} uses a temporal batch size of 1, whereas VideoMAE V2~\cite{wang2023videomae} uses 2), we first resize or repeat the input so that the number of output tokens matches that of our model. For these methods, we generally adopt the “base’’ model checkpoints provided in their official GitHub repositories, as they roughly match the computational budget of our model.

\section{Additional Details on Curriculum Ablation}
\label{sec:supp_curriculum_details}

In this section, we provide additional details on the baseline setups used in Tab.~\ref{tab:ablation_curriculum}. We compare our visual-overlap-based curricula to two baseline strategies:
(1) Non-curriculum baseline, where we do not progressively increase the difficulty of training samples. Concretely, the geometric visual-overlap score remains fixed within the range [0.5, 1.0] throughout training, without any linear decay. As a result, the model encounters challenging samples (\eg, wide-baseline views) from the very beginning.
(2) Frame-interval-based curriculum, where geometric-overlap scores are converted into frame intervals that linearly increase over training. To construct the interval schedule for each dataset, we pre-sample 10K sequences with geometric-overlap scores in [0.5, 1.0] and set the maximum frame interval to the 95th percentile of these sequences. This heuristic implicitly defines dataset-specific hyperparameters that would otherwise need to be \textit{manually tuned}. 

\section{A Pose-supervised Baseline}
\label{sec:supp_pose_supervised_baseline}


We introduce a pose-supervised baseline whose pose estimation module is trained using ground-truth camera poses (typically obtained from running Structure-from-Motion systems~\cite{schonberger2016structure}), following prior supervised methods (\eg, DUSt3R~\cite{wang2023DUSt3R} and VGGT~\cite{wang2025vggt}). In this baseline, the Gaussian-based scene reconstruction module is still optimized with a photometric loss; however, gradients from this loss are not propagated back to the pose estimation module. The results are shown in Tab.~\ref{tab:supp_comparison_pose_supervised}.

We observe that while the pose-supervised baseline usually outperforms \ours{} on coarse pose accuracy (RPA@15°/30°), it consistently achieves lower PSNR for novel-view synthesis. We attribute this weaker NVS performance to a misalignment between the predicted poses and the Gaussian prediction. To supervise pose estimation, the ground-truth camera poses are normalized to a predefined scale (\eg, 1.0), and the pose estimation module learns to predict camera poses at this scale. However, the Gaussian prediction module does not necessarily follow the same scale. In practice, we observe many training instances where the rendered Gaussians fall outside the image plane, providing little or no useful photometric supervision.

In contrast, with our curriculum design, \ours{} learns pose estimation and Gaussian prediction jointly, allowing both components to automatically align to the same scale. This avoids the scale-misalignment issue and leads to more stable training and stronger novel-view synthesis performance. In short, this experiment further confirms the benefit of our self-supervised 3D reconstruction framework for both camera pose estimation and novel-view synthesis.

\section{Additional Results on Pre-training}
\label{sec:supp_pretraining}

We present additional results where \ours{} is used as a pre-trained backbone for VGGT* (our re-implementation of VGGT~\cite{wang2025vggt}, matched to our architecture and training data). We compare \ours{} against RayZer~\cite{jiang2025rayzer} as an alternative pre-training approach and evaluate pose accuracy across multiple datasets.

Tab.~\ref{tab:supp_comparison_with_vggt} summarizes results under two training configurations: using only DL3DV~\cite{ling2024dl3dv} and using a mixture of seven datasets. Note that pre-training and supervised finetuning are conducted on the same data (\ie, DL3DV or the 7-dataset mixture). In both settings, VGGT* initialized with \ours{} outperforms its RayZer-initialized counterpart on most metrics, indicating that the representations learned by \ours{} provide stronger and more transferable pre-training for downstream supervised pose estimation.

\section{Further Analysis of Training Data}
\label{sec:supp_training_data}

We further analyze how different training datasets affect model performance.

Compared to Tab.~\ref{tab:data_mixing_ablation}, Tab.~\ref{tab:supp_data_mixing_ablation} additionally includes \ours{} results on a static subset of SpatialVID~\cite{wang2025spatialvid}, a large in-the-wild video dataset, and reports the number of training sequences used in each setting. We observe that a larger number of training sequences does not necessarily yield higher performance. For example, the model trained on 100K SpatialVID sequences performs comparably to the RealEstate10K~\cite{zhou2018stereo} model (which uses 66K sequences), yet significantly underperforms the DL3DV~\cite{ling2024dl3dv} model (which contains only 10K sequences). We conjecture that this gap stems from the noisy nature of in-the-wild data: SpatialVID sequences originate primarily from internet videos, and our training subsets are selected using their coarse dynamic-ratio labels. Also, SpatialVID often features simple or near-static camera motions. In contrast, DL3DV is carefully curated without moving objects and contains high-quality video sequences with diverse camera trajectories. These results support our earlier observations about data quality and highlight the importance of data curation when scaling self-supervised learning to large in-the-wild resources.

We also find that mixing datasets improves distribution coverage and leads to better generalization. For instance, models trained with mixed data perform better on the object-centric CO3Dv2~\cite{reizenstein2021common} compared to models trained solely on non-object-centric datasets. 

Finally, we note that all experiments are conducted under a fixed computation budget (\ie, 152K iterations with a global batch size of 192). Within this controlled setting, our results consistently suggest that diversity and quality of data matter more than quantity for training self-supervised models. We believe that collecting diverse, high-quality data remains both a key challenge and a promising direction for future work.

\section{More Qualitative Comparisons}
\label{sec:supp_qualitative_comparison}

\noindent \textbf{Learned Feature Representations.} In Fig.~\ref{fig:supp_vis_feature}, we provide additional qualitative results comparing the learned feature representations of \ours{} with those of RayZer~\cite{jiang2025rayzer}. The feature maps produced by \ours{} exhibit more stable and coherent patterns across views, while RayZer’s feature maps often display noticeable color shifts between frames. These results suggest that \ours{} learns feature representations that are more geometrically grounded.

\begin{figure*}[t]
  \centering
  \includegraphics[width=\textwidth]{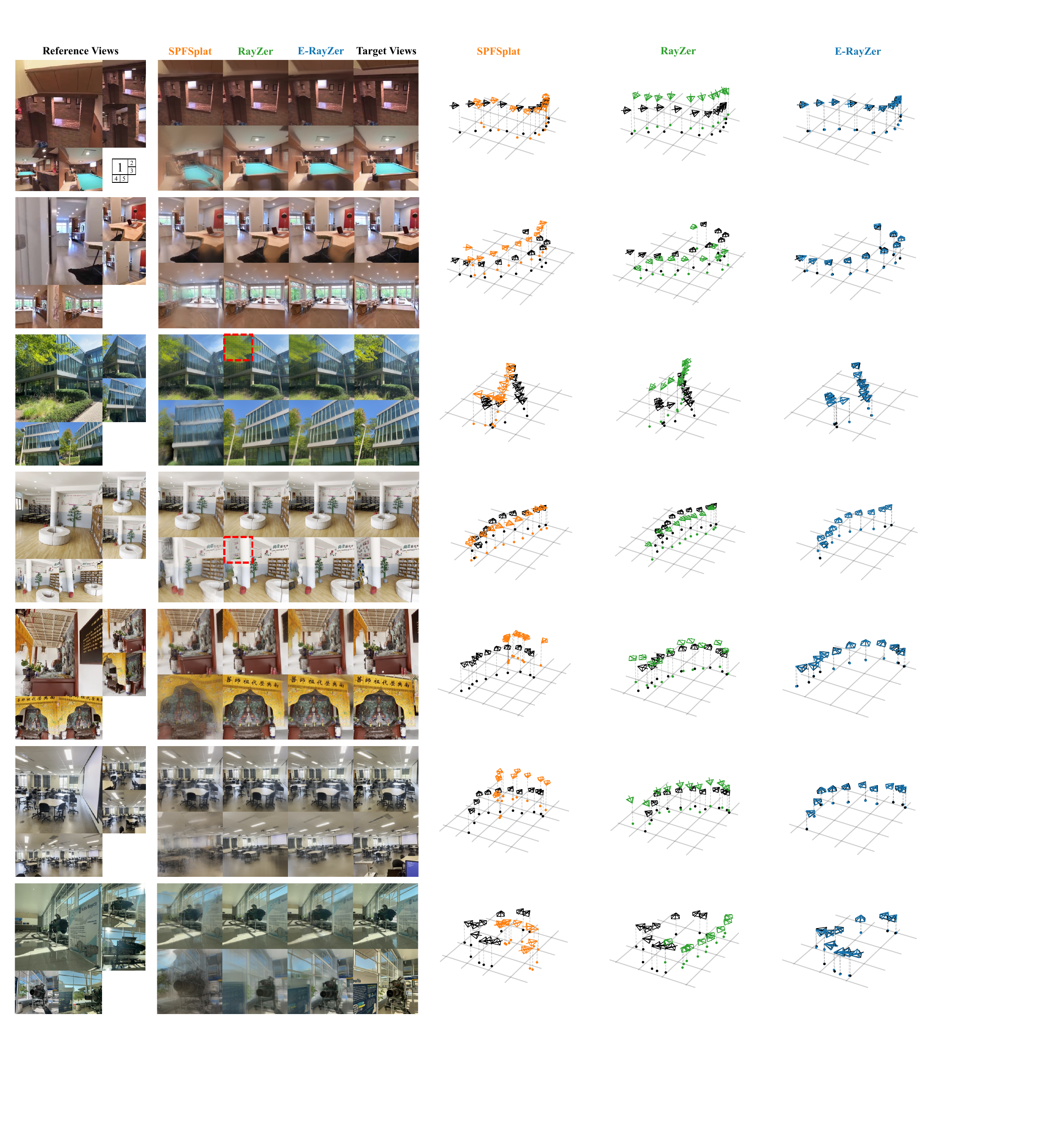}
  \vspace{-0.27in}
  \caption{\textbf{Additional Visual Comparison with (Partially) Self-supervised Methods.} We show results for both novel-view synthesis (left) and pose estimation (right). The temporal order of the reference views is shown in the first row. Ground-truth poses are visualized in black, and predicted poses are aligned to the ground truth via an optimal similarity transform. \ours{} outperforms baselines in pose accuracy, demonstrating its grounded 3D understanding. While RayZer~\cite{jiang2025rayzer} typically produces high-quality novel views, it often exhibits grid-like artifacts in low-texture regions (highlighted with red boxes; best viewed when zoomed in), likely due to its latent-rendering formulation.}
  \label{fig:supp_comparison}
  \vspace{-0.05in}
\end{figure*}

\noindent \textbf{Pose Estimation and Novel-view Synthesis.} We present additional qualitative comparison with baselines in Fig.~\ref{fig:supp_comparison}. Compared to SPFSplat~\cite{huang2025no}, \ours{} consistently achieves better pose accuracy and higher-quality novel-view synthesis, despite being trained entirely from scratch without relying on pretrained priors such as MASt3R~\cite{leroy2024grounding}.
RayZer~\cite{jiang2025rayzer} generally produces high-quality novel views; however, it often exhibits grid-like artifacts in uncertain regions (highlighted with red bounding boxes). 
Moreover, RayZer’s predicted poses are not physically aligned with the scene, whereas the camera poses learned by \ours{} are geometrically grounded.

{
    \small
    \bibliographystyle{ieeenat_fullname}
    \bibliography{main}
}

\end{document}